\documentclass[lettersize,journal]{IEEEtran}
\usepackage{amsmath,amsfonts}
\usepackage{algorithm}
\usepackage{algpseudocode}
\usepackage{array}
\usepackage{xcolor}
\usepackage[caption=false,font=normalsize,labelfont=sf,textfont=sf]{subfig}
\usepackage{textcomp}
\usepackage{stfloats}
\usepackage{url}
\usepackage{verbatim}
\usepackage{graphicx}
\usepackage{multirow,makecell}
\usepackage[nocompress]{cite}
\usepackage{overpic}
\usepackage{graphicx}
\usepackage{pifont}
\usepackage{pict2e} 
\usepackage{booktabs}
\usepackage{multirow}
\usepackage{multicol}
\usepackage{xcolor,colortbl}
\usepackage{nicematrix}

\usepackage{subcaption}
\usepackage[colorlinks,citecolor=blue,linkcolor=blue]{hyperref}

\usepackage{orcidlink}
\usepackage{overpic}
\usepackage{colortbl,color}
\usepackage[table]{xcolor}
\DeclareMathOperator*{\argmax}{arg\,max}

\newcommand{\x}{\mathbf{x}}

\usepackage{physics}
\usepackage{bm}

\begin{document}

\title{Quality-Preserving Imperceptible Adversarial Attack on Skeleton-based Human Action Recognition}

\author{Ziyi Chang$^{\orcidlink{0000-0003-0746-6826}}$, Kanglei Zhou$^{\orcidlink{0000-0002-4660-581X}}$, Xiaohui Liang$^{\orcidlink{0000-0001-6351-2538}}$, Hubert P. H. Shum$^{\orcidlink{0000-0001-5651-6039}}$, ~\IEEEmembership{Senior Member, ~IEEE}
\thanks{Manuscript received \today. This research is supported in part by the EPSRC NortHFutures project (ref: EP/X031012/1). (\textit{Corresponding author: Hubert P. H. Shum.})}
\thanks{Ziyi Chang and Hubert P. H. Shum are with the Department of Computer Science, Durham University, Durham, DH1 4FL, UK (e-mail: \{ziyi.chang, hubert.shum\}@durham.ac.uk).}
\thanks{
Kanglei Zhou is with the Department of Psychology and Cognitive Science, Tsinghua University, Beijing 100084, China (e-mail: zhoukanglei@tsinghua.edu.cn).
}
\thanks{
Xiaohui Liang is with the State Key Laboratory of Virtual Reality Technology and Systems, Beihang University, Beijing 100191, China, and also with the Zhongguancun Laboratory, Beijing 100190, China (e-mail: liang\_xiaohui@buaa.edu.cn).
}
}

\markboth{IEEE Transactions on Circuits and Systems for Video Technology}%
{Chang \MakeLowercase{\textit{et al.}}: Quality-Preserving Imperceptible Adversarial Attack}

\IEEEpubid{
\begin{minipage}{\textwidth}
\vspace{0.3cm}
\centering
Copyright \copyright\ 2026 IEEE. Personal use of this material is permitted.
However, permission to use this material for any other purposes must be obtained
from the IEEE by sending an email to pubs-permissions@ieee.org.
\end{minipage}
}

\maketitle

\begin{abstract}
Adversarial attacks on skeletal human action recognition have received significant attention.
However, existing methods typically introduce noise-like perturbations that degrade motion quality post-attack, and thereby are inherently perceptible with recent advancements in S-HAR systems.
We discover that this degradation stems from the gap between empirical and true risks during the optimization process of previous adversarial attacks.
To address this issue, we propose an attack where adversarial motions are obtained without compromising their motion quality.
To minimize the risk gap and preserve motion quality, we propose a distribution-based adversarial attack method without introducing noise-like perturbations.
To faithfully evaluate the motion quality, we propose a new metric that aligns with human perception on real-world naturalness.
Experiments have been conducted on the state-of-the-art S-HAR methods across two datasets, demonstrating the superiority of our method in both the attack success rate and the post-attack motion quality through qualitative and quantitative analyses.
The success of our quality-preserving attack application and distribution-based method raises serious concerns about the robustness of action recognizers, highlighting the need for further enhancements in this domain.
\end{abstract}


\section{Introduction}

\IEEEPARstart{A}{dversarial} attacks on skeletal human action recognition (S-HAR) have received significant attention due to concerns regarding the robustness of S-HAR systems. These systems have been deployed to recognize human actions from skeletal inputs and have been deployed in various security-critical and human-centric domains such as medical care, action assessment, and safety surveillance \cite{ren2024survey}. The adversarial motions pose substantial threats to such life-critical applications by misleading the S-HAR systems \cite{diao2024tasar}. Therefore, obtaining adversarial motions is important to enhance the robustness of these S-HAR systems and boost trustworthy and responsible artificial intelligence \cite{bai2021recent,chander2024toward}.

Previous adversarial attack methods \cite{wang2021understanding,liu2020adversarial} typically introduce noise-like perturbations to skeletal inputs to deceive S-HAR algorithms, resulting in reduced post-attack motion quality.
These perturbations inherently conflict with human perceptions of natural motion \cite{troje2002decomposing,shimada2012modulation} due to the quality decline, and thereby diminish the imperceptibility of existing attacks.
Nonetheless, this vulnerability is often overlooked due to inherent dataset noise \cite{shahroudy2016ntu,kay2017kinetics,liu2019ntu} and limited classifier expressiveness \cite{madry2018towards,ilyas2019adversarial,li2024adversarial}.
Furthermore, existing metrics that compare paired pre-attack and post-attack motions inadequately assess post-attack motion quality because motions are sparsely distributed and their neighborhoods may not be natural and plausible \cite{karunratanakul2024optimizing}.
With advancements enabling high-quality skeletal inputs for S-HAR systems \cite{kovacs2024lidpose,xu2024finepose,shin2024wham}, the attacked motions become increasingly perceptible and undermine the imperceptibility of adversarial attack.

\begin{figure}[htbp]
\centering
\includegraphics[width=\linewidth]{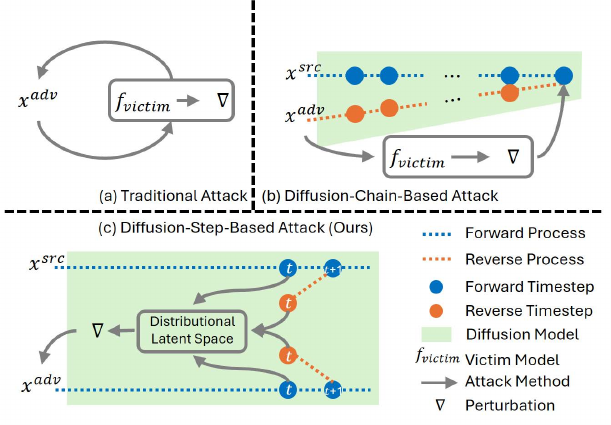}
\caption{The schema of our proposed method when compared with existing attack methods. Instead of relying on the diffusion chain and the gradient from external victims, our method constructs a distributional latent space to leverage the inherent distribution knowledge learned by diffusion models for attack.}
\label{fig:schema}
\end{figure}

\IEEEpubidadjcol

We identify the decline in the quality of adversarial motions resulting from the optimization process in previous attack methods as shown in Fig. \ref{fig:schema}. Firstly, existing methods typically rely on paired loss to calculate perturbations where the empirical risk for optimization significantly deviates from the true risk. The empirical risk typically refers to the loss defined over a large number of observations, while the true risk is defined as the expected loss across the implicit data distribution \cite{vapnik1991principles}. However, in the adversarial attack context, only an individual motion is involved for optimizing a motion to be adversarial, which leads to a large gap between the empirical and true risks \cite{devroye2013probabilistic}. Secondly, the optimization process relies heavily on the gradient signals from a classifier, whose decision boundaries often extend beyond the true data manifold. As a result, the adversarial modification is driven in directions that do not necessarily align with the underlying distribution of natural motions. This classifier-dependent gradient guidance frequently pushes the motion off the manifold, leading to unnatural, noise-like patterns \cite{claeskens2008model} that ultimately degrade the quality and plausibility of the adversarial motions.

\begin{figure*}
\sf
\centering
\begin{overpic}[width=\textwidth,trim=0 0 0 0,clip]{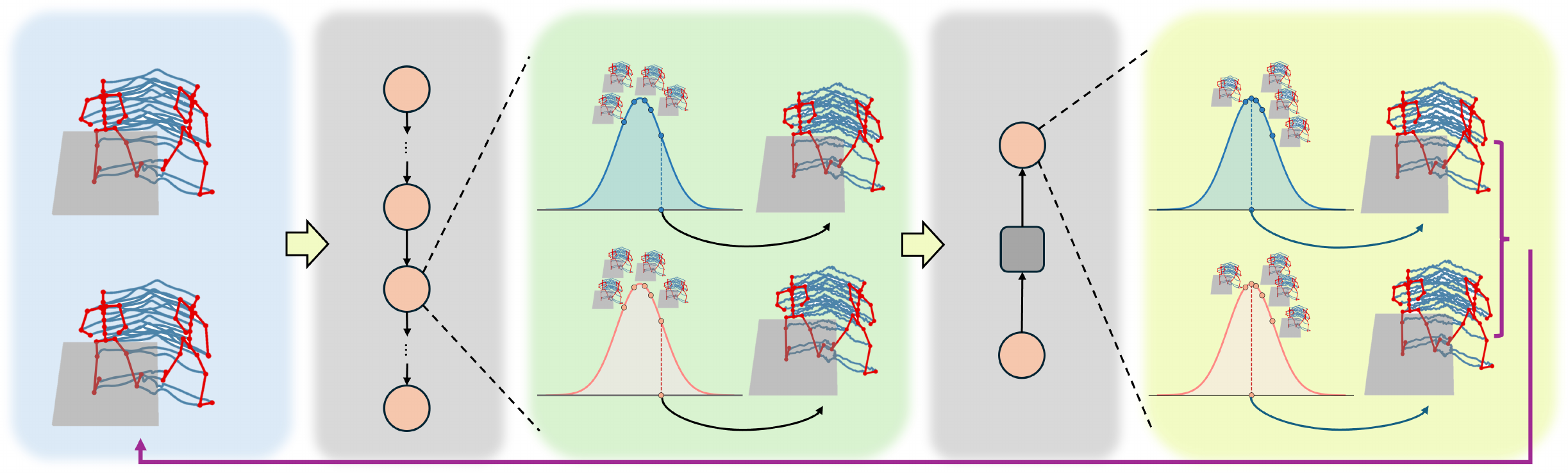}
\put(6,27.4){\footnotesize Data Space}
\put(12,16){$\mathbf{x}^{\mathrm{src}}_0$}
\put(12,2){$\mathbf{x}^{\mathrm{adv}}_0$}
\put(20,28){\footnotesize Diffusion Forward}
\put(20.2,26.4){\scriptsize (Multiple Timesteps)}
\put(25.5,24){$\scriptstyle 1$}
\put(25.6,16.5){$\scriptstyle t$}
\put(24.8,11.4){$\scriptscriptstyle t+1$}
\put(25.4,3.8){$\scriptstyle  T$}
\put(38,27.4){\footnotesize Distributional Latent Space}
\put(54,16){$\mathbf{x}^{\mathrm{src}}_{t+1}$}
\put(54,4){$\mathbf{x}^{\mathrm{adv}}_{t+1}$}
\put(59.5,28){\footnotesize Diffusion Reverse}
\put(60.2,26.4){\scriptsize (Single Timestep)}
\put(64.9,20.3){$\scriptstyle t$}
\put(64.8,13.8){$\scriptstyle\textcolor{white}{\theta}$}
\put(64.1,7.2){$\scriptscriptstyle  t+1$}
\put(82,27.4){\footnotesize Attack Strategy}
\put(92,16){$\bm\mu^{src}_{t}$}
\put(92,4){$\bm\mu^{adv}_{t}$}
\put(96.3,14.5){$\scriptstyle\mathrm{grad}$}
\end{overpic}
\caption{Overview of the proposed distribution-based imperceivable adversarial S-HAR attack where post-attack motion quality is preserved. We optimize the motions along the gradient provided by a pretrained diffusion model based on the distributional latent space to keep the post-attack motion quality.
}
\label{fig:overview}
\end{figure*}
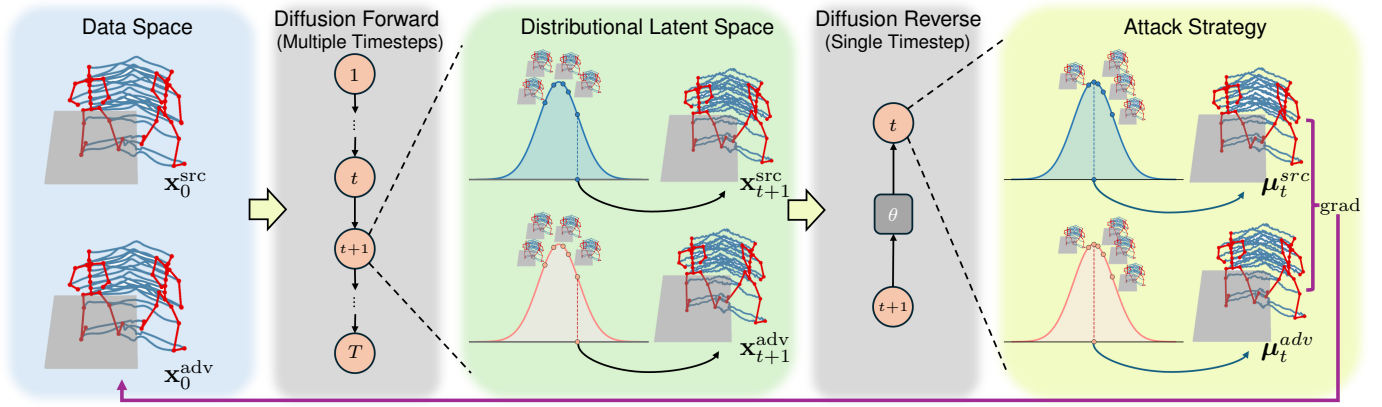

To tackle the decline in the quality of adversarial motions, we propose a novel attack application where adversarial motions are obtained without sacrificing quality.
This application demonstrates a new task towards imperceivable adversarial attack where the post-attack motion quality is explicitly constrained rather than the paired-motion constraints in previous methods.
To minimize the risk gap in the optimization process of adversarial attack, we propose a distribution-based S-HAR attack method where a diffusion model is employed to learn the data distribution{\color{black}, as shown in Fig.\ref{fig:overview}}.
The proposed distribution-based method relies on the data distribution instead of paired pre-attack and post-attack motions to optimize the given motions.
By reducing the risk gap, our optimization does not introduce noise-like perturbations and thereby our adversarial motions are obtained without sacrificing the motion quality.
Furthermore, to faithfully quantify the quality of adversarial motions, we propose a new metric to measure the naturalness based on physiological analysis of human movements \cite{ueyama2021costs,ivanenko2004five}.
This new metric measures the inherent physiological naturalness rather than relying on the paired individuals comparison in existing adversarial metrics, aligning with the human perceptions of natural motions in the real world.

Extensive experiments show that our method generates adversarial motions with the least quality decrease for attacking against the four latest S-HAR classifiers across both a high-quality dataset, i.e., 100STYLE \cite{mason2022local} and two commonly used datasets, i.e., HDM05 \cite{muller2007hdm05} and NTU60 \cite{shahroudy2016ntu} when compared with state-of-the-art S-HAR adversarial attack methods. Our user study further validates that our adversarial motions are the least perceivable by humans. We also conduct ablation studies on hyper-parameters of diffusion models and the specification of generative models. Codes are available in \url{https://github.com/mrzzy2021/QualityPreservingAttack}. Our contributions are summarized as follows:

\begin{itemize}
    \item We discover a critical yet previously overlooked vulnerability in existing attack methods where their noise-like adversarial perturbations are inherently perceptible to humans. To address this, we propose a novel attack framework that imperceptible adversarial attack is fulfilled without sacrificing post-attack motion quality.
    \item To preserve post-attack motion quality, we propose a distribution-based attack method where imperceptible adversarial motions are generated through a pre-trained diffusion model by minimizing the gap between empirical and true risks in our optimization of adversarial attack.
    \item To faithfully quantify the quality of adversarial motions, we propose a new metric based on existing physiological analysis of real-world human motions, enabling the assessment of previously overlooked vulnerabilities that existing metrics fail to capture.
\end{itemize}

The remainder of this paper is organized as follows: We first review related work in Section \ref{sec:related_work}. Then, we formulate the proposed quality-preserving adversarial attack application in Section \ref{sec:application_formulation}. Subsequently, we present our proposed distribution-based adversarial attack method with quality preservation in Section \ref{sec:distribution_method}. We formulate the proposed metric in Section \ref{sec:quality_metric} to faithfully evaluate the motion quality after attack. Finally, we demonstrate the superiority of our method in Section \ref{sec:experiment_result} and provide a summary in Section \ref{sec:conclusion_discussion} with potential future work.

\section{Related Work}\label{sec:related_work}

\subsection{General Adversarial Attack}
Adversarial attacks are first introduced in \cite{szegedy2013intriguing}, highlighting the vulnerability of deep neural networks and subsequently extending to various data types.
Generally, adversarial attacks serve as a specialized form of data augmentation aimed at exposing system vulnerabilities by generating new samples, whereas other data augmentation techniques may focus on objectives such as enhancing training efficiency and improving inference performance \cite{shanmugam2021better}.
Deep neural networks remain susceptible to meticulously crafted adversarial attacks despite that significant achievements have been witnessed. Noise-like perturbations have been applied to input data to easily deceive these high-performing neural networks and people raise concerns on the trustworthiness and reliability of deployed neural networks \cite{bai2021recent,chander2024toward}. In response to these concerns, researchers have extensively explored adversarial attacks across different data modalities, including 2D images \cite{xie2025retouchuaa,gan2025lesep}, videos \cite{videopure2025jiang,huang2025transhfc}, 3D objects \cite{wang2025unified}, physical objects \cite{athalye2018synthesizing}, and graph data \cite{li2020deeprobust}. While adversarial research on different modalities has been widely investigated, attacks on time-series data have recently gained attention \cite{karim2020adversarial}, with relatively less focus on 3D skeletal motions characterized by spatio-temporal structures.

\subsection{Adversarial Attacks on S-HAR}
Adversarial attacks on skeletal human action recognition (S-HAR) aim to perturb 3D human skeletons to deceive classifiers, mirroring objectives from the image domain.
Skeletal motion is extensively utilized in HAR to mitigate challenges such as lighting variations, occlusions, and diverse view angles \cite{ren2024survey}.
Consequently, the vulnerability of skeleton-based classifiers to adversarial attacks has received increasing attention.
With a significantly lower degree of freedom in motions \cite{lu2023hard,diao2024understanding}, previous S-HAR attack methods focus on improving imperceptibility through different perspectives.
\cite{liu2020adversarial} targets GCN-based models by employing generative adversarial networks as a discriminator to achieve imperceptibility in terms of the anthropomorphic plausibility.
{\color{black}\cite{tanaka2022adversarial,cao2025bones} propose a new attack method where only the length of bones could be perturbed to enhance the imperceptibility from the perspective of bone length consistency.}
\cite{wang2021understanding} analyzes the perceptibility of adversarial skeletal samples and proposes constraints between paired pre-attack and post-attack motions, integrating the observations from \cite{liu2020adversarial} and \cite{tanaka2022adversarial} to further improve imperceptibility. {\color{black}\cite{kang2023qesar} proposes to find perturbation directions on the manifold of motions to pursue imperceptibility.}
However, while these methods effectively achieve to fool the S-HAR system, they inherently introduce noise-like perturbations that degrade post-attack motion quality, and thereby undermine the imperceptibility of post-attack motions.

\subsection{Diffusion Models} 
Diffusion models \cite{sohl2015deep} have attracted extensive attention for their powerful generative capabilities.
These models utilize numerous timesteps to transform data distributions into a standard Gaussian distribution through a forward process, while a denoising network is trained to recover the less noisy data distribution \cite{chang2023design}.
A large number of timesteps are typically required for smooth distribution transitions and simultaneously provide a hierarchical latent space with rich semantics.

Some have leveraged the generative ability of diffusion models for adversarial attack.
\cite{liu2024boosting, chen2023advdiffuser} directly learns adversarial generation using paired adversarial samples, though this dependency may not be practical for real-world applications. \cite{liu2024unstoppable} introduces a label-based generative method, training diffusion models on auxiliary datasets.
\cite{hu2024towards} leverages diffusion models with modified denoising process to generate adversarial samples. \cite{hu2024establishing} utilizes diffusion models for high-quality image steganography. \cite{chen2024diffilter} focuses on purification as a defense approach by the generative ability of diffusion models rather than the adversarial attack. {\color{black}Some rely on DDIM inversion to achieve adversarial attack \cite{gan2025lesep,yang2024diffusion} or defence \cite{jiang2025videopure} through the full denoising chain of diffusion models.}
Despite the applications mentioned above, the previous diffusion-based methods still rely heavily discriminative classifier to calculate the perturbation and do not leverage the inherent distributional knowledge contained by the diffusion model itself.

\section{Imperceptibility of S-HAR Adversarial Attack via Motion Quality Preservation}\label{sec:application_formulation}

In this section, we formulate our proposed attack application, which conducts imperceptible adversarial attack against skeleton-based human action recognition (S-HAR) systems without compromising post-attack motion quality.

\paragraph{Inherent Vulnerability of Imperceptibility}
Existing adversarial attacks achieve imperceptibility by introducing small, noise-like perturbations that inherently and detrimentally affect post-attack motion quality.
Motion quality encompasses naturalness and plausibility \cite{zhu2023human}, as human motions adhere to physical and biomechanical constraints.
The human brain has specialized neural mechanisms for perceiving biological motion \cite{blakemore2001perception,grossman2000brain} and is highly sensitive to unnatural and implausible kinematics \cite{troje2002decomposing,shimada2012modulation}.
Given the highly nonlinear and articulated spatial-temporal structures of human motions \cite{wang2021understanding}, even subtle noise-like perturbations significantly degrade quality.
Such perturbations disrupt motion dynamics and physical constraints, and deviate motions from the natural and plausible distribution. Thus, perturbed motions suffer from the decline in motion quality and become perceptible.

\paragraph{Task Formulation} Nonetheless, the inherent perceptible vulnerability resulting from the introduced noise-like perturbations is often overlooked by previous S-HAR attack methods due to the intrinsic noise in datasets \cite{shahroudy2016ntu,kay2017kinetics,liu2019ntu} and the limited expressiveness capacity of classifiers \cite{madry2018towards,ilyas2019adversarial,li2024adversarial}.
As advancements \cite{kovacs2024lidpose,xu2024finepose,shin2024wham} have widely facilitated cutting-edge S-HAR systems to utilize skeletons with high quality as inputs for decision-making, human supervisors readily perceive suspicious noise-perturbed skeletal inputs.

Furthermore, previous constraints during their optimization process, which usually focus on paired pre-attack and post-attack motions, fail to faithfully reflect the quality of adversarial motions.
Since motions are sparsely distributed \cite{karunratanakul2024optimizing}, attacked motions that remain within the neighborhood of their pre-attack motions usually may not be natural and plausible, especially when considering the motion dynamics.
Therefore, adversarial motions suffering from post-attack quality become increasingly perceptible due to the added noise-like perturbations, which urgently need to be resolved to improve the imperceptibility of adversarial attacks.

To achieve imperceptibility by preserving the quality of adversarial samples instead of introducing noise-like perturbations, we propose an adversarial task, in which post-attack motions preserve pre-attack motion quality.
Specifically, our task explicitly requires the quality of the post-attack motions as the optimization constraint instead of individual motion pairs, and aims to achieve imperceptibility by preserving post-attack motion quality.
We define our adversarial task as follows:

\begin{equation}
    \begin{aligned}
    \underset{\mathbf{x}^\text{adv}}{\text{minimize}} \quad & \mathcal{L}_{\text{adv}}(C(\mathbf{x}^{\text{adv}})) \\
    \text{subject to} \quad & \mathcal{Q}(\mathbf{x}^{\text{adv}})=\mathcal{Q}(\mathbf{X}^{\text{src}}),
    \end{aligned}
\end{equation}
where $C(\cdot)$ is an S-HAR classifier, $\mathbf{x}^\text{adv}\in\mathbb{R}^{T\times J\times3}$ represents the attacked motion, $\mathbf{X}^\text{src}\in\mathbb{R}^{N\times T\times J\times3}$ represents the set of natural and plausible motions, and $Q(\cdot)$ represents the motion quality.
Compared with previous adversarial attacks, we propose optimizing an individual motion sample, $\mathbf{x}^\text{adv}$, with respect to the entire set of natural and plausible motions rather than rely on a pair of pre-attack and post-attack motions.

\section{Distribution-based S-HAR Attack Method}\label{sec:distribution_method}

In this section, we demonstrate the proposed distribution-based adversarial attack method against skeleton-based human action recognition (S-HAR) systems.
First, we identify the origin of noise-like perturbations introduced in existing attack methods and propose integrating the data distribution into adversarial attacks by constructing a generative diffusion distributional latent, as described in Section \ref{sec:diffusion_latent}.
Then, we present our attack strategy to leverage the constructed latent for a quality-preserving adversarial attack in Section \ref{sec:attack_strategy}.

\subsection{The Diffusion Latent for Quality Preservation}\label{sec:diffusion_latent}

\paragraph{The Risk Gap of Previous Optimization} 
The noise-like perturbations in previous attacks are attributed to the gap between empirical and true risks in their optimization processes.
Empirical risk typically refers to the loss defined over a large number of observations, while true risk is defined as the expected loss across the implicit data distribution \cite{vapnik1991principles}.
However, in previous adversarial attacks, the optimization process involves only an individual motion sample, leading to a large gap between empirical and true risks \cite{devroye2013probabilistic}.
When optimizing a motion with respect to a single pre-attack motion observation, noise-like perturbations are usually introduced \cite{claeskens2008model} and decrease the quality of adversarial motions.

\paragraph{Constructing Generative Diffusion Latent}

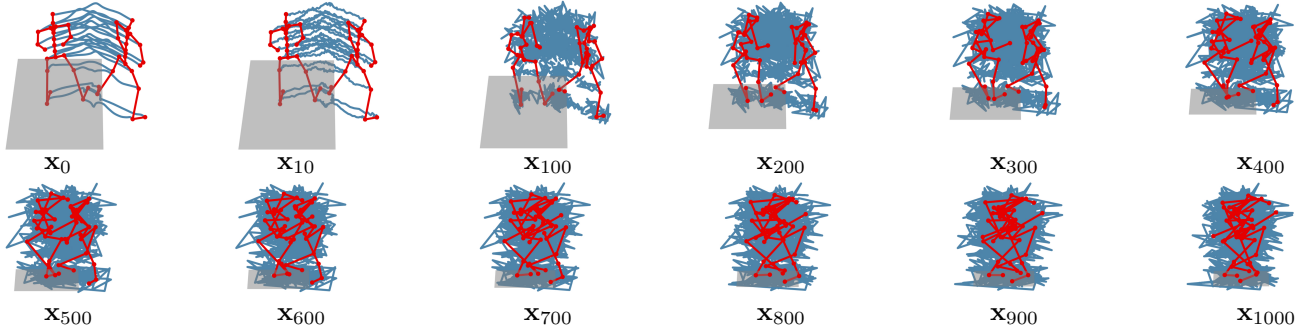
\begin{figure*}
\sf
\centering
\begin{overpic}[width=\textwidth,trim=0 0 0 0,clip]{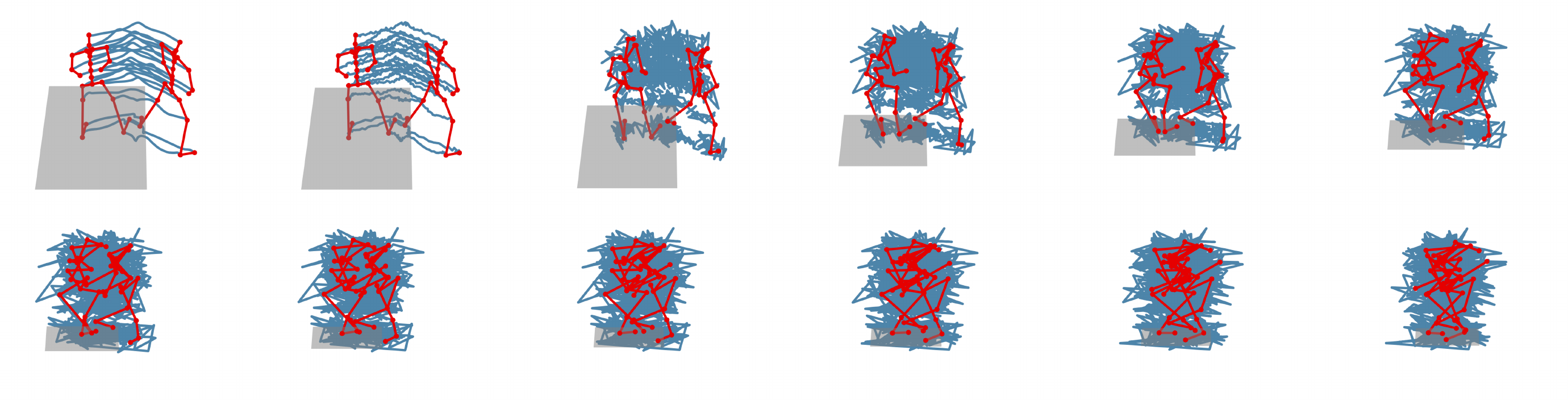}
\put(5,12){$\mathbf{x}_0$}
\put(5,1){$\mathbf{x}_{500}$}
\put(22,12){$\mathbf{x}_{10}$}
\put(22,1){$\mathbf{x}_{600}$}
\put(40,12){$\mathbf{x}_{100}$}
\put(40,1){$\mathbf{x}_{700}$}
\put(57,12){$\mathbf{x}_{200}$}
\put(57,1){$\mathbf{x}_{800}$}
\put(74,12){$\mathbf{x}_{300}$}
\put(74,1){$\mathbf{x}_{900}$}
\put(92,12){$\mathbf{x}_{400}$}
\put(92,1){$\mathbf{x}_{1000}$}
\end{overpic}
\caption{The visualization of diffusion latents at different timesteps. As shown, the earlier timesteps maintain more low-level details, and the later timesteps focus on high-level structures until latents become pure noise.}
\label{fig:latentvis}
\end{figure*}

To minimize the risk gap in the optimization process of adversarial attacks, we propose leveraging distributional latents from generative diffusion models for our quality-maintaining adversarial attack.
As shown in Fig. \ref{fig:latentvis}, diffusion models have semantically meaningful latent spaces that capture different feature patterns with respect to various choices of timesteps \cite{wu2023latent}. When compared to discriminative models \cite{jaini2024intriguing}, they are capable of modelling motion distributions that are critical for quality preservation. Compared with other generative models like VAEs \cite{petrovich2021action}, their hierarchical structures are more comprehensive to represent underlying patterns to modify motions.

Our attack leverages diffusion distributional latents as proxies for the given pre-attack motion individual.
Previous S-HAR adversarial attack methods typically optimize motions directly in the original data space.
However, the original space is sparse where natural motions being sparsely distributed \cite{karunratanakul2024optimizing}.
This sparsity leads to difficulties in modifying motions within the original data space while maintaining their quality.
In contrast, we build our attack method on the stochastic latent space of diffusion models.
The stochastic distributional latents are located in an approximately smooth space \cite{preechakul2022diffusion} and encode different levels of semantics for adversarial modification.
By transforming the given pre-attack motion into latent distributions, the empirical risk of our optimization involves infinite samples within distributions rather than individual motions.

Specifically, we define our latent for adversarial attack to include two parts: a posterior mean for its rich semantics encoded in the stochastic process of diffusion models and a stochastic latent for its regularization.
The posterior mean combines both the forward and reverse processes in a diffusion model, thereby comprehensively and hierarchically representing the motion.
First, we leverage the forward process to convert motions into a distribution, which is defined as:
\begin{equation}\label{eq:forward}
    p(\mathbf{x}_{t+1}|\mathbf{x}_0) = \mathcal{N}(\sqrt{\alpha_{t+1}}\mathbf{x}_0, (1-\alpha_{t+1})\mathbf{I}),
\end{equation}
where $\mathbf{x}_0$ is a pre-attack motion, and $\mathbf{x}_{t+1}$ is a stochastic sample at the timestep $t+1$ from a Gaussian distribution.
Then, to achieve finer-grained latent manipulation for adversarial attack, we propose to construct our latent proxy for adversarial attack in a single timestep of diffusion reverse process.
The posterior distribution in the reverse process is defined as:
\begin{align}
\label{eq:forward_process_posterior}
    p(\mathbf{x}_{t}|\mathbf{x}_{t+1}, \mathbf{x}_0) = \mathcal{N}(\bm\mu, \sigma_t \mathbf{I}),
\end{align}
where $\mathbf{x}_{t+1}\sim p(\mathbf{x}_{t+1}|\mathbf{x}_0)$, $\sigma_{t}$ is the posterior variance and $\bm\mu$ is the posterior mean of the distribution. Since the posterior variance is pre-defined in noise schedule, $\bm\mu$ represents the denoising direction according to the class label.
To show the relationship between the posterior mean and data distribution, we explicitly derive the posterior mean based on Tweedie’s formula \cite{kim2021noise2score} and formulate it as:
\begin{equation}
    \bm\mu_t=\gamma_t\mathbf{x}_0+\lambda_t\epsilon_t+\delta_t\nabla\log p_\theta(\mathbf{x}_{t+1},\mathbf{y}),
\end{equation}
where $\mathbf{x}_0$ is the motion in original data space, $\mathbf{x}_{t+1}$ is sampled from the latent distribution, i.e., $\mathbf{x}_{t+1}\sim p(\mathbf{x}_{t+1}|\mathbf{x}_0)$ and $\epsilon_t\sim \mathcal{N}(0,\mathbf{I})$. The posterior mean is a latent that encapsulates the prior knowledge of the entire dataset's distribution. The first term, $\mathbf{x}_0$ ensures content preservation during the attack, while the distributional term is responsible for aligning the data distribution to remain post-attack motion quality.

Different from the previous methods that regularize the attack process using point-wise difference, i.e., with respect to $\mathbf{x}_0$, we add the regulation over the latent $\mathbf{x}_t\sim p(x_t|x_0)$ to regularize the data distribution during attack.

In summary, the motions during attack are represented by
\begin{equation}\label{eq:posterior_mean}
    \bm\kappa_t = \bm\mu_t+\bm x_t,
\end{equation}
{\color{black}which will be leveraged later in the attack process as discussed in Section \ref{sec:attack_strategy}.}

\paragraph{Benefits of Distributional Latent}
As shown in Eq. \ref{eq:posterior_mean}, the constructed latent, i.e., the posterior mean with regularization, is derived from the data distribution rather than from a single sample.
Compared with deterministic latents \cite{song2021denoising}, the stochastic latents are more efficient and have much higher capacity \cite{wu2023latent}. Moreover, the posterior mean represents a direction that conforms to the data distribution toward high-density areas associated with the given labels.
When we optimize over this latent for the adversarial attack, our empirical risk is based on all data samples from the approximated distribution, instead of a single sample.
Additionally, instead of computing over the entire diffusion chain, we focus on only one denoising step to minimize changes in a single optimization iteration.
Through sampling different timesteps, the constructed latent captures different features that encode underlying semantics.

\subsection{The Attack Strategy for Quality-Preserving Attack}\label{sec:attack_strategy}
\paragraph{Overview of Strategy}
To leverage the diffusion latent space for adversarial attacks, we propose perturbing the source motion observation by randomly sampling different class labels rather than relying on the gradient of a specific classifier.
This is not only because obtaining the gradient of a specific classifier is challenging in real-world applications \cite{diao2024tasar}, but also because the gradient of a classifier may not be reliable for preserving motion quality.
As classifiers focus on the label distribution rather than data distribution, their gradient may point to out-of-distribution regions. While following this gradient achieves the shortest trajectory of deceiving classifiers, it leads to a decline in post-attack motion quality and undermines the imperceptibility.
As a result, we design our method to only rely on the classifier's decisions to determine when to stop and drive the motion towards adversarial samples by sampling adversarial labels for the conditional diffusion model.

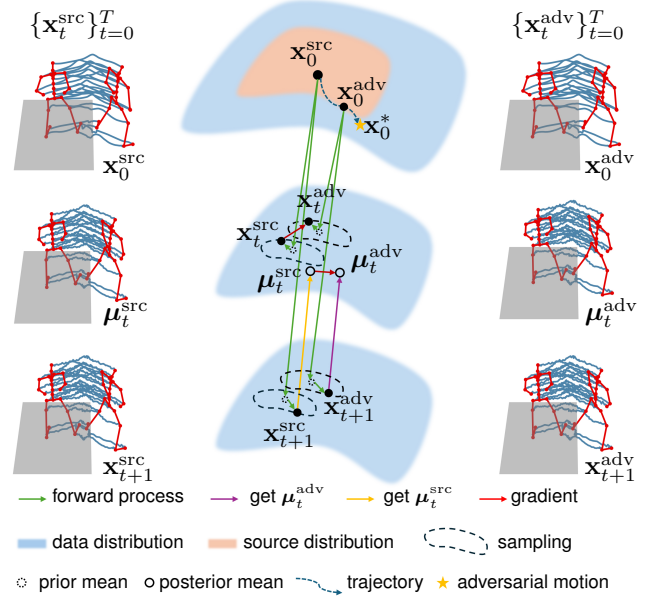
\begin{figure}[t]
\sf
\centering
\begin{overpic}[width=\linewidth,trim=0 0 0 0,clip,]{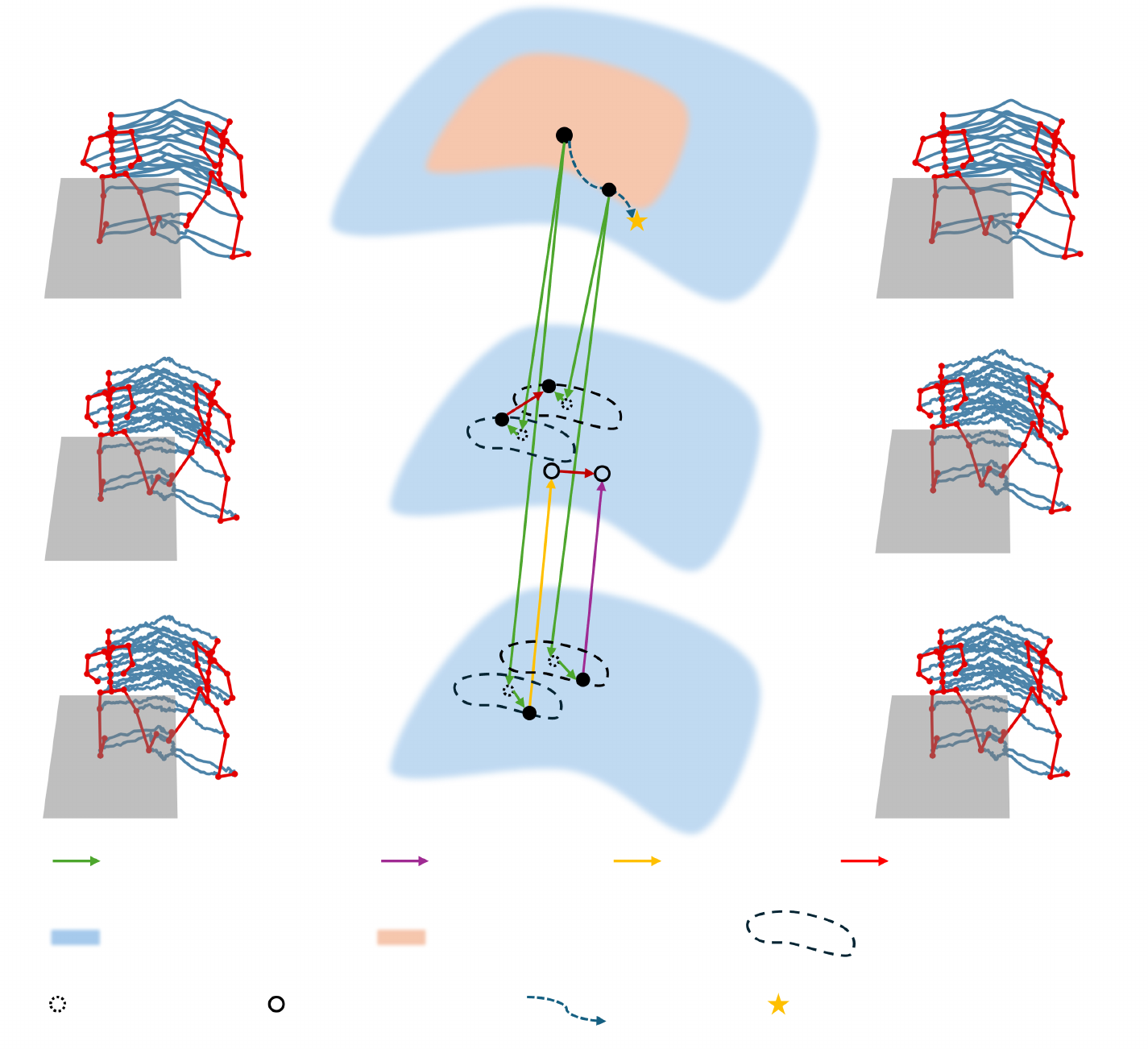}
\put(6,86){$\{\mathbf{x}^\mathrm{src}_t\}_{t=0}^T$}
\put(78,86){$\{\mathbf{x}^\mathrm{adv}_t\}_{t=0}^T$}
\put(17,65){$\mathbf{x}^\mathrm{src}_0$}
\put(89,65){$\mathbf{x}^\mathrm{adv}_0$}
\put(45,82){$\mathbf{x}^\mathrm{src}_0$}
\put(52,76){$\mathbf{x}^\mathrm{adv}_0$}
\put(56,71){$\mathbf{x}^\mathrm{*}_0$}

\put(17,43){$\bm\mu^\mathrm{src}_t$}
\put(89,43){$\bm\mu^\mathrm{adv}_t$}
\put(40,48){$\bm\mu^\mathrm{src}_t$}
\put(54,51){$\bm\mu^\mathrm{adv}_t$}

\put(17,20){$\mathbf{x}^\mathrm{src}_{t+1}$}
\put(89,20){$\mathbf{x}^\mathrm{adv}_{t+1}$}
\put(41,25){$\mathbf{x}^\mathrm{src}_{t+1}$}
\put(50,29){$\mathbf{x}^\mathrm{adv}_{t+1}$}

\put(37,55){$\mathbf{x}^\mathrm{src}_t$}
\put(46,60){$\mathbf{x}^\mathrm{adv}_t$}

\put(9.5,15.8){\scriptsize forward process}
\put(39,15.8){\scriptsize get $\displaystyle\bm\mu^\mathrm{adv}_t$}
\put(59,15.8){\scriptsize get $\displaystyle\bm\mu^\mathrm{src}_t$}
\put(78,15.8){\scriptsize gradient}
\put(9.5,9){\scriptsize data distribution}
\put(38,9){\scriptsize source distribution}
\put(76,9){\scriptsize sampling}
\put(7.5,3){\scriptsize prior mean}
\put(25.5,3){\scriptsize posterior mean}
\put(53.5,3){\scriptsize trajectory}
\put(70,3){\scriptsize adversarial motion}

\end{overpic}
\caption{The illustration of the attack strategy. We illustrate an intermediate calculation at the timestep $t$ during the optimization of achieving the final adversarial motion $\mathbf{x}_0^*$.}
\label{fig:method}
\end{figure}

\paragraph{Optimization Objective} Specifically, we denote the stochastic distributional latents of the source motion observation and the adversarial motion as $\bm\kappa_t^\text{src}$ and $\bm\kappa_t^\text{adv}$. The source motion and the adversarial motion are mapped to the latent space by the diffusion forward process and then the desired posterior mean are obtained within a single timestep denoising conditioned on the ground truth label $\mathbf{y}^\text{src}$ and the randomly sampled the adversarial label $\mathbf{y}^\text{adv}$ from the set of all possible labels excluding the ground truth label, respectively. The regularizer is obtained through the forward process. We define our objective function as follows:
\begin{equation}
    \begin{aligned}
    \mathcal{L}_{\bm\kappa_t}:=&
    0.5\times\mathbb{E}_{t, \epsilon_{t}}\left[
    \Vert \bm\kappa_{t}^\text{adv} - \bm\kappa_{t}^\text{src} \Vert_2^2\right],\\
    =&0.5\times\mathbb{E}_{t, \epsilon_{t}}
    \Vert \underbrace{\bm\mu_{t}^\text{adv} - \bm\mu_{t}^\text{src}}_{\text{latent}} + \underbrace{(\bm\x_{t}^\text{adv} - \bm\x_{t}^\text{src})}_{\text{regularization}} \Vert_2^2
\end{aligned}
\end{equation}
where $\bm\kappa^\text{src}_t$ and $\bm\kappa^\text{adv}_t$ are our defined latents and serve as proxies for adversarial attack. The first term, $\bm\kappa^\text{adv}_0$, represents the direction towards being adversarial, while the second term, {\color{black}$\bm\kappa^\text{src}_0$}, represents the direction of maintaining the representative semantics within the original class. Our optimization objective is an expectation over the diffusion timesteps $t$ and the stochastic distributional latent, indicated by the randomly sampled noise $\epsilon$ in the forward process where semantics are encoded into the latent space of the diffusion model.

\begin{algorithm}[t]
\caption{Diffusion-based Quality-Preserving Adversarial Motion Attack on S-HAR}\label{alg:algorithm}
\begin{algorithmic}[1]
\Require Diffusion model $\theta$, a classifier $\varphi$, a motion $\mathbf{x}_0^\text{src}$ with label $\mathbf{y}\in \mathbf{Y}$, maximum iteration $I$, diffusion timesteps $T$
\State $\mathbf{x}_0^\text{adv} \gets \mathbf{x}_0^\text{src}$
\State $i \gets 0$
\State $\mathbf{y}^\text{adv}\sim \mathbf{Y}_{/\ \mathbf{y}^\text{src}}$ \Comment{Randomly sample an adversarial label}
\While{$i \leq I~and~\mathbf{y}^\text{pred}\neq \mathbf{y}^\text{src}$}
\State $t\sim [1, T]$
\State $\kappa_t^\text{src}=\mu_t(\mathbf{x}_0^\text{src},\mathbf{y}^\text{src};\theta) + \mathbf{x}_t^\text{src}$ \Comment{Eq. \ref{eq:posterior_mean}}
\State $\kappa_t^\text{adv}=\mu_t(\mathbf{x}_0^\text{adv},\mathbf{y}^\text{adv};\theta) + \mathbf{x}_t^\text{adv}$ \Comment{Eq. \ref{eq:posterior_mean}}
\State $\mathrm{grad}=\kappa_t^\text{adv}-\kappa_t^\text{src}$\Comment{Eq. \ref{eq:grad}}
\State $\mathbf{x}^\text{adv}=\mathbf{x}^\text{adv}+\mathrm{grad}$
\State $\mathbf{y}^\text{pred}=\argmax p_\varphi(\mathbf{y}|\mathbf{x}^\text{adv})$ \Comment{Get classifier decision}
\EndWhile
\end{algorithmic}
\end{algorithm}

By minimizing the defined object, our optimization aligns the stochastic distributional latents of the source and the adversarial motions. Specifically, we calculate the gradient of $\mathcal{L}_{\kappa_t}$ with respect to $\bm\kappa^\text{adv}_t$ and obtain our adversarial gradient:
\begin{equation}
\begin{aligned}
\label{eq:grad}
    \mathrm{grad}:=&\nabla\mathcal{L}_{\kappa_t}\\
    =&\mathbb{E}_{t, \epsilon_t}\left[\kappa_{t}^\text{adv} - \kappa_{t}^\text{src}\right],
\end{aligned}
\end{equation}
on which we rely to iteratively update the $\mathbf{x}_0^{\text{adv}}$. The Eq.~\ref{eq:grad} facilitates the adversarial effect and quality preservation. Instead of directly requiring the paired pre-attack and post-attack motions to be close to each other in data space, our optimization aligns the latent distribution of pre-attack and post-attack motions. Constraining the two distributional latents $\bm\kappa_t^\text{adv}$ and $\bm\kappa_t^\text{src}$ promotes that the latent distributions of $\mathbf{x}_0^\text{adv}$ and $\mathbf{x}_0^\text{src}$ remain closely aligned rather than merely examine the pre-attack and the post-attack motions and rely on the gradients from an external classifier. Our optimization represents a single timestep examination, which is different from a multi-timestep generative process conditioned on $\mathbf{y}^\text{adv}$. By optimizing over the expectation of randomly sampled timesteps, the trajectory defined by the posteriors is expected to be closely aligned with each other. Consequently, our attack strategy facilitates the generation of $\mathbf{x}_0^\text{adv}$ that corresponds with $\mathbf{y}^\text{adv}$ to deceive the target model, while simultaneously preserving the quality of $\mathbf{x}_0^\text{src}$. Detailed adversarial attack methodology is provided in Algorithm \ref{alg:algorithm} and Fig.~\ref{fig:method}.

\paragraph{Relationship with Previous Optimization}
We further demonstrate that our proposed method implicitly integrates previous approaches while additionally offer the advantage of distributional prior knowledge provided by a pre-trained diffusion model to minimize the risk gap.
The Eq.~\ref{eq:grad} is equivalently represented using the input motion and the learned distribution, from which the following detailed formulation is derived:
\begin{equation}
    \begin{aligned}
    \label{eq:final_pds_form}
        \mathrm{grad} &:=\mathbb{E}_{t, \epsilon_t}\left[\kappa_{t}^\text{adv} - \kappa_{t}^\text{src}\right]\\
         &~= \mathbb{E}_{t, \epsilon_t} \Bigg[
                \psi(t)(\underbrace{\mathbf{x}_0^{\text{adv}} - \mathbf{x}_0^{\text{src}}}_{\substack{\text{Distance Function}\\\text{in Original Space}}}) \\
            & \quad + \chi(t) \Bigg((\underbrace{\nabla \log p(\mathbf{x}_t^\text{adv})-\nabla \log p(\mathbf{x}_t^\text{src})}_{\text{Distribution Constraint}})\\
            & \quad + (\underbrace{\nabla \log p(\mathbf{y}^\text{adv}|\mathbf{x}_t^\text{adv})}_{\substack{\text{Adversarial}\\\text{Gradient}}}-\underbrace{\nabla \log p(\mathbf{y}^\text{src}|\mathbf{x}_t^\text{src})}_{\substack{\text{Representiveness}\\\text{of the Given Input}}}) \Bigg)
        \Bigg],
    \end{aligned}
\end{equation}
where we decompose the probability with Bayes' theorem, {\color{black}$\psi(t)=\sqrt{\alpha_{t+1}}$, and $\chi(t)=1-\alpha_{t+1}$}. The four terms serve distinct yet complementary purposes in our objective function.
The first term demonstrates a similar measurement to previous attack methods, ensuring that the adversarial sample does not deviate excessively from the input sample.
However, our ability of quality preservation, which distinguishes our method from previous methods, is ensured through the remaining terms.
Unlike traditional adversarial optimization, which considers only an individual motion pair as shown in the first term, our perturbation strategy provides the modification gradient based on data distributions that are composed of infinite motions.
The second term measures the distribution density and ensures that the adversarial motions remain not only neighborhood but also in high-density areas.
The third term considers the adversarial gradient while the fourth term involves the representativeness of the observed sample as a consideration for the range of pre-attack motion neighborhood by evaluating how representative the given input is with respective to the label.
These considerations leverage the distributional prior knowledge and enables the quality preservation under adversarial modifications.

\section{Human Perception Aligned Naturalness Metric}\label{sec:quality_metric}

In this section, we examine previous metrics and formulate the proposed metric to faithfully measure the quality of generated adversarial motions in terms of naturalness.

\paragraph{Unfaithfulness of Existing Metrics}
The evaluation of existing metrics fails to accurately measure quality. Previous metrics rely on paired comparisons between the pre-attack and post-attack motions to assess quality, especially naturalness \cite{wang2021understanding}. However, remaining within the neighborhood of a pre-attack motion does not ensure motion quality comparable to clean motions because natural and plausible motions are sparsely distributed \cite{karunratanakul2024optimizing}. Consequently, these metrics cannot reliably determine whether adversarial motions are sufficiently natural due to their misalignment with human perception.

\paragraph{Human Perception Aligned Metric}
Natural human motions are subject to biomechanical constraints where movements are driven by muscle activation. Every action results from the brain sending electrical signals to nerves, which in turn contract muscles to move joints \cite{chiquier2023muscles}. The human brain develops prior knowledge of natural motions based on real-world observations of biomechanically validated musculoskeletal movements \cite{schneider2024mint} rather than paired comparison. Therefore, human perception is inherently biased toward the naturalness self-embedded in the observed motion.

To faithfully reflect naturalness, we propose a novel metric that evaluates the physiological naturalness of generated adversarial motions. Since real-world movements are created by muscle activation, our metric is grounded in the bio-activity of muscles. Specifically, the muscle activation patterns of natural human motions are constrained by physiological characteristics \cite{ueyama2021costs,ivanenko2004five}, resulting in smooth activation curves. We introduce a new metric to measure quality from the biomechanical perspective of muscle activation:
\begin{equation}
    \mathrm{naturalness}=\frac{1}{T}\int_{0}^{T} \frac{\mathrm{d}^4 J(t)}{\mathrm{d}t^4}\mathrm{d}t,
\end{equation}
where $T$ is the temporal length of a motion, $J(t)$ is the joint position at time $t$. {\color{black} BASAR \cite{diao2021basar} mainly enforces naturalness as source-relative manifold plausibility, whereas our work targets distribution-level quality preservation and evaluates motion naturalness as an intrinsic biomechanical property of the attacked motion itself.}

\section{Experiments}\label{sec:experiment_result}

In this section, we begin by outlining the experimental settings. Subsequently, we quantitatively and qualitatively analyze the performance of our proposed method and existing adversarial techniques. Finally, we perform a user study to empirically evaluate the imperceptibility of post-attack motions and also ablation studies to validate our configurations.

\begin{table*}[]
\caption{Generated Adversarial Motion Quality Comparison on 100STYLE dataset \cite{mason2022local}.}
\centering
\begin{tabular}{cccccccc}
\toprule
Victim & Method & Success Rate $\uparrow$ & FID $\downarrow$ & MMD $\downarrow$ & Physiological Naturalness $\downarrow$ & Foot Skating $\downarrow$ & Bone Variation $\downarrow$ \\ 
\toprule
\multirow{5}{*}{Style \cite{zhong2025smoodi}} 
                          & I-FGSM & 81.72\% & 194.95 & 0.053 & 129.42 & 0.147 & 10.96 \\
                          & MI-FGSM& 82.08\% & 195.22 & 0.053 & 131.33 & 0.147 & 11.16 \\
                          & MIG    & 70.61\% & 238.23 & 0.071 & 264.46 & 0.234 & 23.22 \\
                          & SMART  & 42.65\% & 242.12 & 0.072 & 97.73 & 0.119 & 4.00 \\
                          & Ours   & \textbf{100\%} & \textbf{18.91} & \textbf{0.011} & \textbf{16.05}  & \textbf{0.075} & \textbf{2.94}  \\ \midrule
\multirow{5}{*}{STTFormer \cite{qiu2022spatio}}
                          & I-FGSM & 80.29\% & 162.55 & 0.043 & 190.98 & 0.196 & 19.28 \\
                          & MI-FGSM& 80.29\% & 162.78 & 0.043 & 191.10 & 0.197 & 19.29 \\
                          & MIG    & 72.76\% & 190.75 & 0.050 & 280.56 & 0.241 & 27.63 \\
                          & SMART  & 29.39\% & 195.06 & 0.053 & 103.56 & 0.167 & 3.92 \\
                          & Ours   & \textbf{100\%} & \textbf{20.28} & \textbf{0.013} & \textbf{16.42}  & \textbf{0.077} & \textbf{3.09} \\ \midrule 
\multirow{5}{*}{SkateFormer \cite{do2024skateformer}}
                          & I-FGSM & 68.46\% & 79.63 & 0.019 & 43.90 & 0.053 & 4.53 \\
                          & MI-FGSM& 68.82\% & 79.06 & 0.019 & 44.70 & 0.053 & 4.61 \\
                          & MIG    & 60.22\% & 102.77 & 0.026 & 70.72 & 0.069 & 7.30 \\
                          & SMART  & 31.90\% & 125.46 & 0.030 & 29.32 & \textbf{0.047} & \textbf{1.60} \\
                          & Ours   & \textbf{100\%}   & \textbf{20.16} & \textbf{0.014} & \textbf{16.33}  & 0.079          & 2.99 \\ \midrule
\multirow{5}{*}{FR-Head \cite{zhou2023learning}}
                          & I-FGSM & 86.38\% & 226.80 & 0.068 & 312.90 & 0.276 & 25.65 \\
                          & MI-FGSM& 86.74\% & 226.67 & 0.068 & 316.77 & 0.280 & 26.08 \\
                          & MIG    & 78.14\% & 242.85 & 0.075 & 489.37 & 0.396 & 38.12 \\
                          & SMART  & 32.97\% & 262.75 & 0.100 & 214.63 & 0.301 & 6.33 \\
                          & Ours   & \textbf{100\%}   & \textbf{19.13} & \textbf{0.011} & \textbf{16.75}  & \textbf{0.084} & \textbf{3.23} \\ \midrule 
\multirow{5}{*}{\textit{Average}}
                          & I-FGSM & 79.21\% & 165.98 & 0.046 & 169.3 & 0.168 & 15.11 \\
                          & MI-FGSM& 79.48\% & 165.93 & 0.046 & 170.98 & 0.169 & 15.29 \\
                          & MIG    & 69.34\% & 193.65 & 0.056 & 276.28  & 0.235 & 24.07 \\
                          & SMART  & 34.23\% & 206.35 & 0.064 & 111.31 & 0.159 & 3.96  \\
                          & Ours   & \textbf{100\%}   & \textbf{19.62} & \textbf{0.012} & \textbf{16.39}  & \textbf{0.079} & \textbf{3.06}  \\
                          \bottomrule
\end{tabular}
\label{tab:comparison_100style}
\end{table*}

\subsection{Experimental Settings}

\paragraph{Datasets}
To evaluate our imperceptible adversarial attacks via quality preserving, we select the 100STYLE \cite{mason2022local} dataset due to its noise-free and inherently high-quality characteristics.
The 100STYLE dataset is collected using a motion capture system and comprises 100 classes of different styles. We represent the skeletons using Cartesian coordinates for 23 joints. This dataset is pre-processed by segmenting long sequences into several segments according to the valid periods provided by \cite{mason2022local}. Additionally, we employ the HDM05 \cite{muller2007hdm05} dataset and NTU60 \cite{shahroudy2016ntu} to evaluate our method. HDM05 and NTU60 are commonly used for action recognition tasks with lower quality compared to 100STYLE. We adhere to the pre-processing procedure outlined in \cite{wang2021understanding}. The HDM05 dataset includes 65 classes of different human actions, and the hip joint is fixed to the origin after pre-processing.

\paragraph{Evaluated Models}
Given the significant advancements in the field of human action recognition and defense methods, we adopt the latest S-HAR models as victim classifiers with randomized smoothing \cite{cohen2019certified} as a defense method to effectively evaluate the performance of adversarial attack methods against advanced classifiers. Specifically, we select the Style classifier \cite{zhong2025smoodi}, STTFormer \cite{qiu2022spatio}, Skateformer \cite{do2024skateformer}, and FR-Head \cite{zhou2023learning} as victim models. These models encompass both transformer-based and graph-based architectures. We utilize their publicly available codebases to train their models.

\paragraph{Evaluation Metrics} Human motions are governed by physical and biomechanical constraints, thereby necessitating the assessment of visual motion quality in terms of naturalness and plausibility \cite{zhu2023human}. To evaluate the motion quality in adversarial attacks, we introduce the physiological naturalness metric, as discussed in Section \ref{sec:quality_metric}, which is grounded in the physiological characteristics of natural human movements. Additionally, we report the Frechet Inception Distance (FID) and Maximum Mean Discrepancy (MMD) based on acceleration to measure the naturalness by assessing the distributional similarity between pre-attack and post-attack motions. As for plausibility, we report the foot skating ratio and bone length variations between frames. Following \cite{guo2022generating}, foot skating is quantified by the consistency between foot velocity and foot height. A high foot skating ratio indicates a significant violation in terms of physical constraints. Bone length variation \cite{duan2024physics} measures the consistency of bone lengths across frames. Higher deviations indicate distortions in the skeleton structure. Finally, we report the success rate of the adversarial attacks to evaluate the threatfulness.

\paragraph{Attacking Methods}
We compare our method with the state-of-the-art (SOTA) S-HAR attack technique, i.e. SMART \cite{wang2021understanding}, as well as other adversarial attack methods including I-FGSM \cite{kurakin2018adversarial2}, MI-FGSM \cite{dong2018boosting}, and MIG \cite{ma2023transferable}. To ensure a fair comparison, we execute 2000 iterations for each attack method, allowing all methods to explore a broader solution space in their pursuit of effective adversarial motions. Since our method requires a pre-trained diffusion model, we adopt the diffusion model proposed by \cite{tevet2023human} and follow the prescribed settings to pre-train it on the two datasets.

\begin{table*}[]
    \caption{Generated Adversarial Motion Quality Comparison on HDM05 dataset.}
    \centering
    \begin{tabular}{ccccccc}
    \toprule
    Victim & Method & Success Rate $\uparrow$ & FID $\downarrow$ & MMD $\downarrow$ & Physiological Naturalness $\downarrow$ & Bone Variation $\downarrow$ \\
    \toprule
    \multirow{5}{*}{Style \cite{zhong2025smoodi}}
                              & I-FGSM  & 93.91\% & 129.78 & 0.086 & 1226.33 & 126.39 \\
                              & MI-FGSM & 93.91\% & 129.79 & 0.086 & 1226.98 & 126.46 \\
                              & MIG     & 92.47\% & 194.33 & 0.170 & 2273.09 & 238.34 \\
                              & SMART   & 68.46\% & 142.39 & 0.116 & 1052.44 & 41.00 \\
                              & Ours    & \textbf{100\%}   & \textbf{22.91} & \textbf{0.012} & \textbf{187.28} & \textbf{32.15} \\ \midrule
    \multirow{5}{*}{STTFormer \cite{qiu2022spatio}}
                              & I-FGSM  & 86.74\% & 162.98 & 0.124 & 1488.08 & 144.97 \\
                              & MI-FGSM & 86.74\% & 163.18 & 0.125 & 1487.92 & 144.95 \\
                              & MIG     & 76.34\% & 202.24 & 0.188 & 1945.83 & 189.95 \\
                              & SMART   & 75.63\% & 174.86 & 0.129 & 974.33 & 33.30 \\
                              & Ours  & \textbf{100\%}   & \textbf{21.89} & \textbf{0.011} & \textbf{185.73} &               \textbf{30.31} \\ \midrule
    \multirow{5}{*}{Skateformer \cite{do2024skateformer}}
                              & I-FGSM  & 83.87\% & 36.66 & 0.022 & 521.52 & 54.50 \\
                              & MI-FGSM & 84.23\% & 36.83 & 0.022 & 523.93 & 54.83 \\
                              & MIG     & 78.14\% & 57.59 & 0.037 & 692.53 & 78.87 \\
                              & SMART   & 65.59\% & 50.84 & 0.028 & 364.26 & \textbf{13.54} \\
                              & Ours    & \textbf{100\%}   & \textbf{18.67} & \textbf{0.011} & \textbf{181.89} & 28.09 \\ \midrule
    \multirow{5}{*}{FR-Head \cite{zhou2023learning}}
                              & I-FGSM  & 96.06\% & 169.34 & 0.124 & 1902.39 & 177.91 \\
                              & MI-FGSM & 96.77\% & 169.96 & 0.126 & 1920.98 & 179.56 \\
                              & MIG     & 92.47\% & 215.88 & 0.206 & 2757.84 & 243.96 \\
                              & SMART   & 81.72\% & 211.88 & 0.173 & 1574.95 & 55.46 \\
                              & Ours    & \textbf{99.28\%} & \textbf{20.80} & \textbf{0.013} & \textbf{182.96} & \textbf{31.07}\\ \midrule
    \multirow{5}{*}{\textit{Average}}  
                              & I-FGSM  & 90.15\% & 124.69 & 0.089 & 1284.58 & 125.94 \\
                              & MI-FGSM & 90.41\% & 124.94 & 0.090 & 1289.95 & 126.45 \\
                              & MIG     & 84.86\% & 167.51 & 0.150 & 1917.32 & 187.78 \\
                              & SMART   & 72.85\% & 144.99 & 0.112 & 991.50 & 35.83 \\
                              & Ours    & \textbf{99.82\%} & \textbf{21.07} & \textbf{0.012} & \textbf{184.47} & \textbf{30.41} \\
                              \bottomrule
    \end{tabular}
    \label{tab:comparison_hdm05}
\end{table*}

\begin{table*}[]
    \caption{Generated Adversarial Motion Quality Comparison on NTU60 dataset.}
    \centering
    \begin{tabular}{ccccccc}
    \toprule
    Method & Success Rate $\uparrow$ & FID $\downarrow$ & MMD $\downarrow$ & Physiological Naturalness $\downarrow$ & Bone Variation $\downarrow$ \\ 
    \toprule
    I-FGSM  & 95.07\% & 3.73 & \textbf{0.002} & 398.99 & 40.33 \\
    MI-FGSM & 95.43\% & \textbf{3.60} & \textbf{0.002} & 400.57 & 40.46 \\
    MIG     & 93.10\% & 8.10 & 0.003 & 412.59 & 44.25 \\
    SMART   & 67.57\% & 432.67 & 0.377 & 403.91 & \textbf{36.94} \\
    Ours    & \textbf{100\%} & 10.50 & 0.005 & \textbf{395.78} & 39.93 \\
    \bottomrule
    \end{tabular}
    \label{tab:comparison_ntu60}
\end{table*}

\subsection{Adversarial Motion Quality Evaluation}

We qualitatively and quantitatively evaluate the performance of adversarial attack in terms of deceitfulness, motion quality including naturalness and plausibility, and human imperceptibility.
Deceitfulness measures the effectiveness of an adversarial method.
In addition to deceitfulness, motion quality assesses whether the adversarial motions are plausible and natural, and serves as an indicator for imperceptibility.
Beyond analytical measurements, we empirically evaluate human imperceptibility. Imperceptibility demonstrates the possibility that humans cannot distinguish pre-attack and post-attack motions when they are mixed together as S-HAR inputs. For simplicity, we present only representative results in this paper.

\paragraph{Deceitfulness}
The success rates presented in Table \ref{tab:comparison_100style} indicate that our method is the most deceitful compared with other methods on the 100STYLE dataset.
Our method consistently achieves an average success rate of $100\%$ even in the face of a randomized smoothing defense strategy. Our attack effectively modifies nearly every input to become adversarial without relying on the gradient of victim model within a limited number of iterations.
In contrast, other methods struggle to generate adversarial motions within iteration constraints.
This demonstrates that the stochastic latent features convey comprehensive semantics \cite{park2023understanding} about the underlying dynamics via different timesteps.

\paragraph{Motion Quality}
Regarding naturalness, the FID and MMD scores in Table \ref{tab:comparison_100style} show that the distribution of our generated motions closely resembles that of natural motions, whereas other methods exhibit significant distributional deviations.
Our adversarial motions maintain proximity to the ground truth distribution of motion dynamics by a considerable margin.
The lowest FID and MMD scores indicate that our adversarial motions are indistinguishable from those in the ground truth dataset.
In other words, the post-attack motion quality is well preserved because our method leverages the distributional knowledge introduced by the pre-trained diffusion model.
By utilizing data distribution, our method minimizes the gap between empirical and true risks by assessing deviations between the modified motions and the data distribution.
Conversely, other methods can only access single source motions, and disrupt the naturalness of post-attack motions, resulting in much higher FID and MMD scores.

\begin{figure*}
\centering
\includegraphics[width=\linewidth]{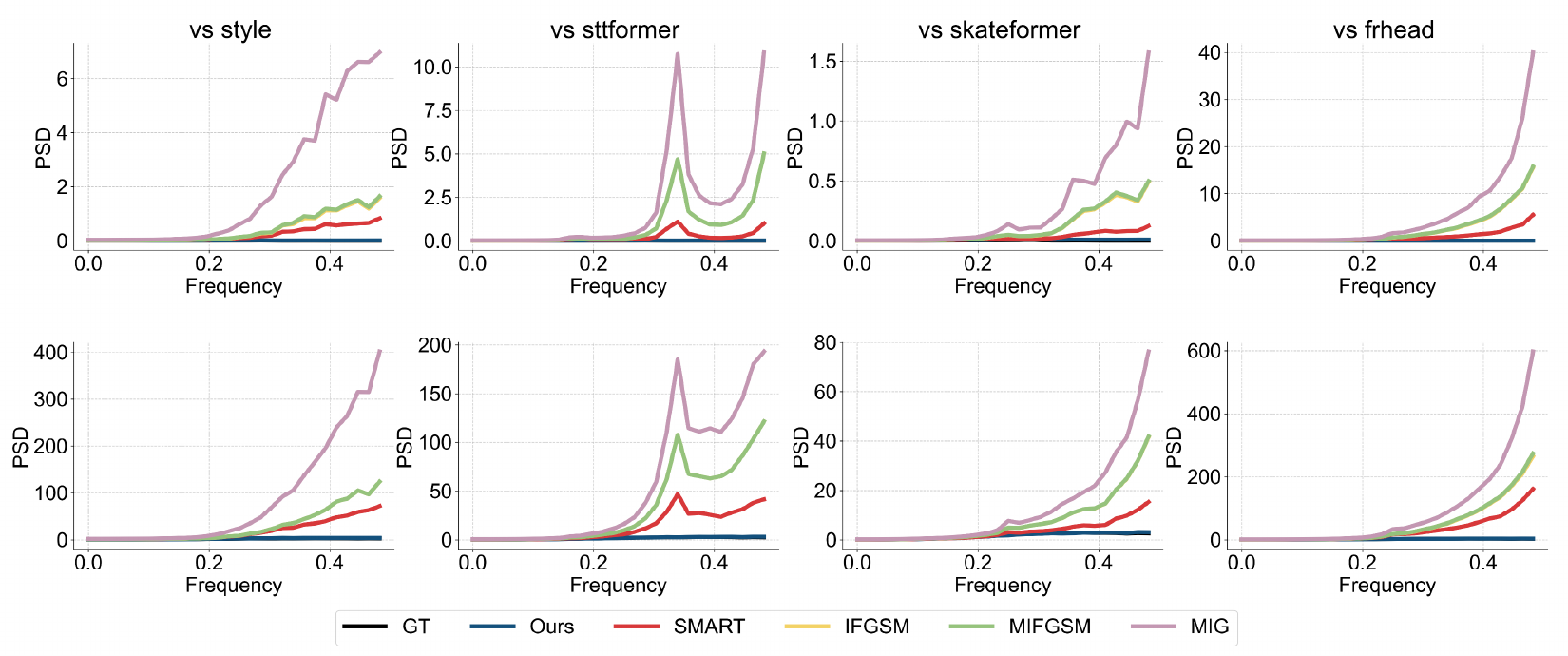}
\caption{The mean power spectral density of adversarial samples found on 100STYLE (upper row) and HDM05 (lower row) against four classifiers. 
}
\label{fig:psd}
\end{figure*}

Physiological naturalness reflects whether adversarial motions conform to natural movements from the biomechanical perspective of muscle activation.
Our method achieves the best performance by a large margin, as shown in Table \ref{tab:comparison_100style}.
This indicates that even after adversarial attacks, our generated motions retain plausible movements that real-world humans can physically perform.
To further evaluate naturalness, as illustrated in Fig.~\ref{fig:psd}, we present the mean power spectral density (PSD) of adversarial samples.
The spectral density of our post-attack motions in the high-frequency domain is significantly lower than that of other adversarial motions. Besides, our spectral density closely aligns with the ground truth curve.
This suggests that our adversarial motions suffer from few noise-like perturbations because we minimize empirical risk over the data distribution rather than a single sample.

\begin{figure}
\centering
\includegraphics[width=\linewidth]{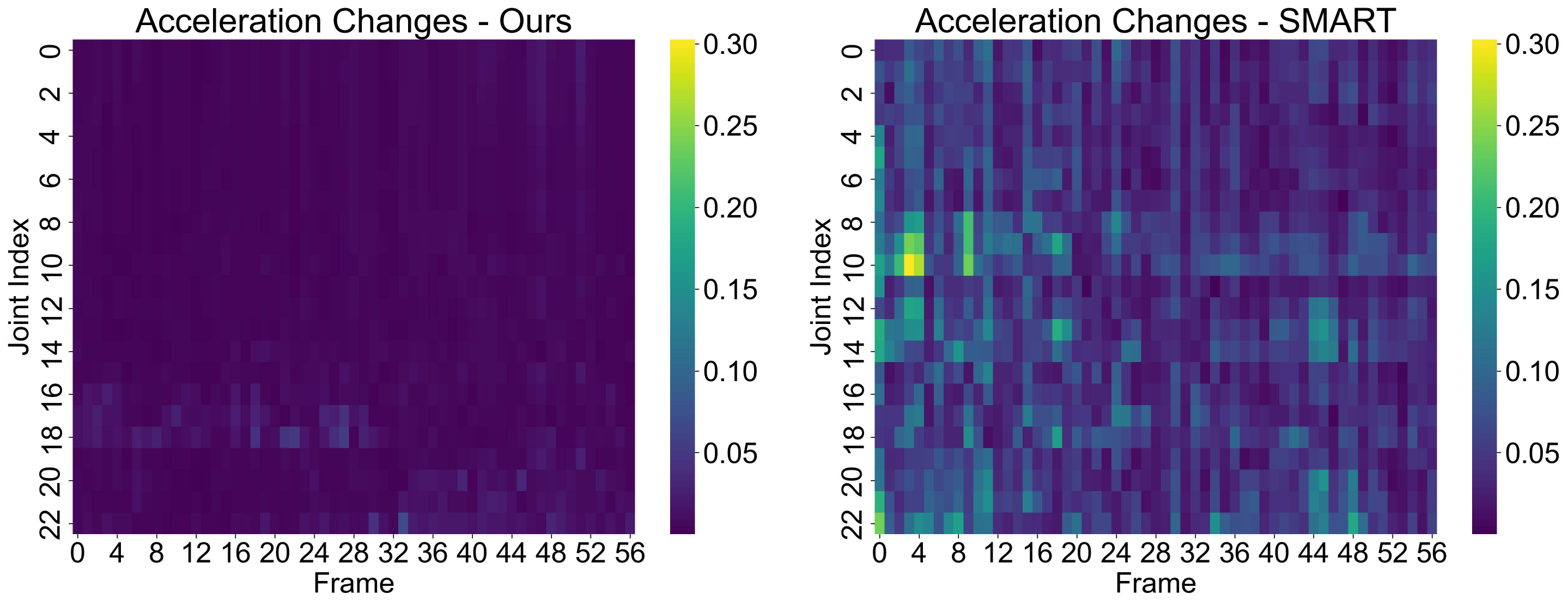}
\caption{The visualization of acceleration changes.}
\label{fig:acc_change}
\end{figure}

We further compare the effect of explicitly regulating motion dynamics with respect to a given sample in SMART and implicitly regulating motion dynamics through data distribution in our diffusion-based method.
Fig.~\ref{fig:acc_change} displays the acceleration changes of a sample for comparison.
Our method introduces modifications that are not only smaller in magnitude but also smoother and more consistent, whereas SMART does not perturb the motion coherently and consistently.
These results imply that minimizing empirical risk over only a given sample leads to deviations from the natural motion distribution and results in the post-attack quality decline.

In terms of physical plausibility, the foot sliding ratio and bone length variation in Table \ref{tab:comparison_100style} demonstrate that our generated adversarial motions exhibit superior movement coherency and skeletal consistency compared to other methods.
The foot skating ratio indicates the coherence of movements concerning foot contact with the ground.
Our method achieves the lowest foot skating ratio, suggesting that post-attack movements are still well synchronized.
This is attributed to our method's ability to minimize the risk gap through the integration of data distribution rather than relying on a single sample.
Additionally, our adversarial motions exhibit the lowest variation in bone lengths, maintaining consistent skeletal structures across frames.
By leveraging access to the data distribution rather than individual samples, our method ensures that modifications preserve cross-frame skeletal integrity.

\begin{figure*}
\scriptsize
\sf
\setlength{\tabcolsep}{0.38em}
\begin{tabular}{m{0.1cm}m{8.7cm}}
\makecell{\begin{tabular}{c}
\\[10pt]
\rotatebox{90}{I-FGSM}      \\[20pt]
\rotatebox{90}{MI-FGSM}  \\[30pt]
\rotatebox{90}{MIG} \\[30pt]
\rotatebox{90}{SMART} \\[25pt]
\rotatebox{90}{Ours} \\[25pt]
\end{tabular}}     & \includegraphics[width=\textwidth,clip]{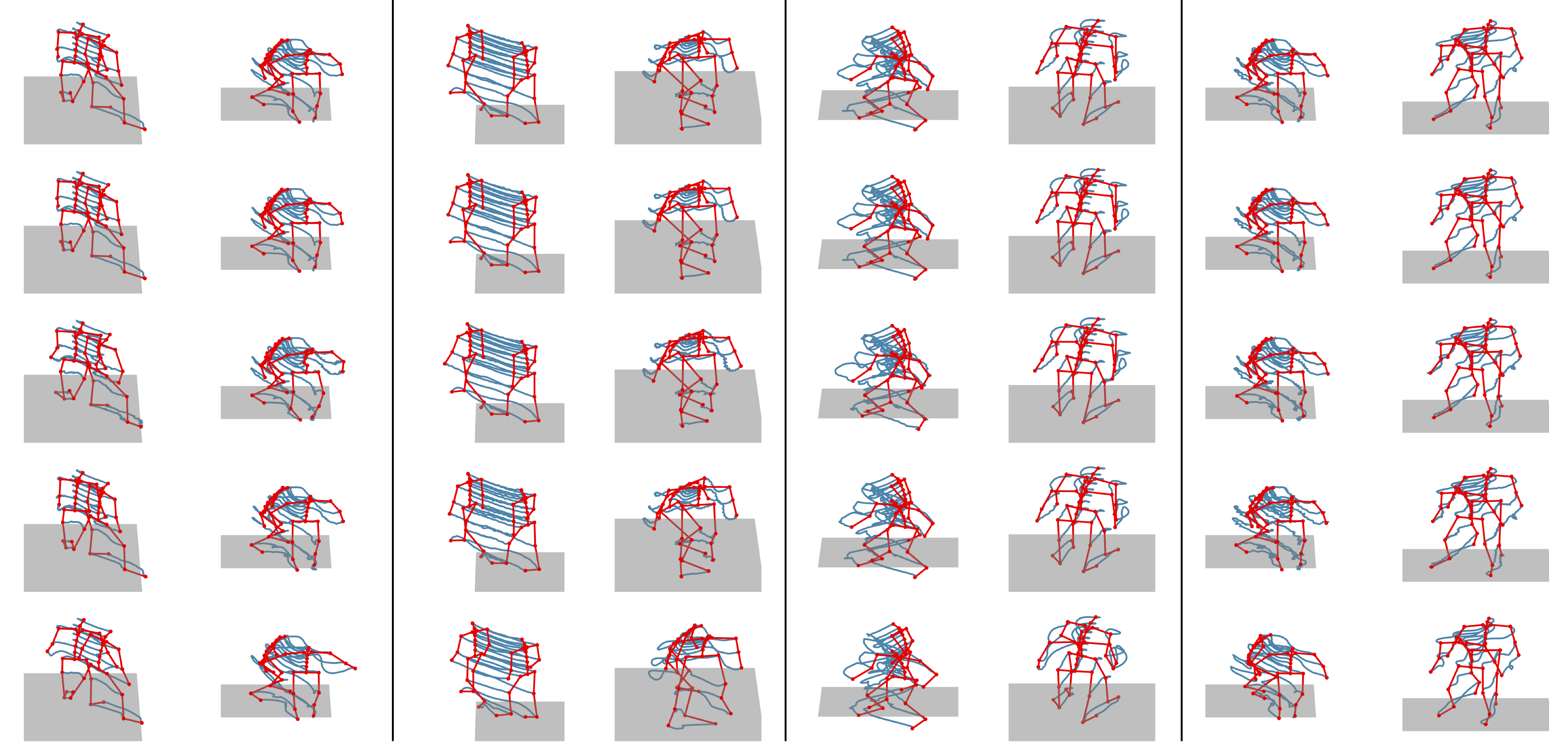} \\
     & \makecell{\begin{tabular}{m{4.4cm}<{\centering}m{4.4cm}<{\centering}m{4.4cm}<{\centering}m{4.4cm}<{\centering}}
       Style \cite{zhong2025smoodi} &  STTFormer \cite{qiu2022spatio} & SkateFormer \cite{do2024skateformer} & FR-Head \cite{zhou2023learning} \\
     \end{tabular}}
\end{tabular}
\caption{Visual comparison among the adversarial motions generated by different attack methods against victim models. We visualize the starting and the ending poses in red, the trajectories of all joints in blue, and the ground floor in grey. Our adversarial motions exhibit the most natural and stable trajectories.} 
\label{fig:qualitative}
\end{figure*}

We also achieve the best performance on the HDM05 dataset, as shown in Table \ref{tab:comparison_hdm05}. 
Our method successfully fools all systems with an average success rate of 99.82\%.
This further validates the efficacy of our attack method even when the data quality and amount are not as high as those of the 100STYLE dataset, demonstrating its threatfulness to such systems.
Moreover, the distribution of our adversarial motions closely aligns with the ground truth dataset, evidenced by the lowest FID and MMD scores. This validates that our adversarial motions are as natural as pre-attack motions. Besides, physiological naturalness indicates that our generated motions adhere to real-world biomechanical constraints.
Since the HDM05 dataset data have been centered to the origin of the coordinates, we do not report the foot skating ratio for this dataset.
Additionally, our method achieves superior motion plausibility with the lowest bone length variation. Both the best physiological and physical plausibility demonstrate that our method preserves the quality of adversarial motions.

The NTU60 dataset has also been tested with a randomized smoothing defence strategy as shown in Table \ref{tab:comparison_ntu60}. Our method still outperforms other methods in terms of success rate and physiological naturalness. However, the motions in this dataset suffer heavily from inherent noise. The adversarial noise introduced by other methods cannot be reflected by the FID and MMD due to the inherent noise from data.

\paragraph{Human Imperceivability}
We visualize the trajectories of joints in adversarial motions generated by our method and previous methods in Fig.~\ref{fig:qualitative} against the four victim models.
In real-world natural movements, joint trajectories are typically smooth and stable. 
As shown in Fig.~\ref{fig:qualitative}, our method produces the most stable and smooth trajectories for all joints in the adversarial motions.
By minimizing the risk gap over the data distribution rather than a single motion, our optimization results do not introduce noise-like perturbations, whereas other methods suffer from significant gaps by optimizing over a single input.
Consequently, the noise-like perturbations lead to a decline in the post-attack motion quality with observable unstable trajectories and undermine imperceptibility.

\begin{figure}
\centering
\includegraphics[width=\linewidth]{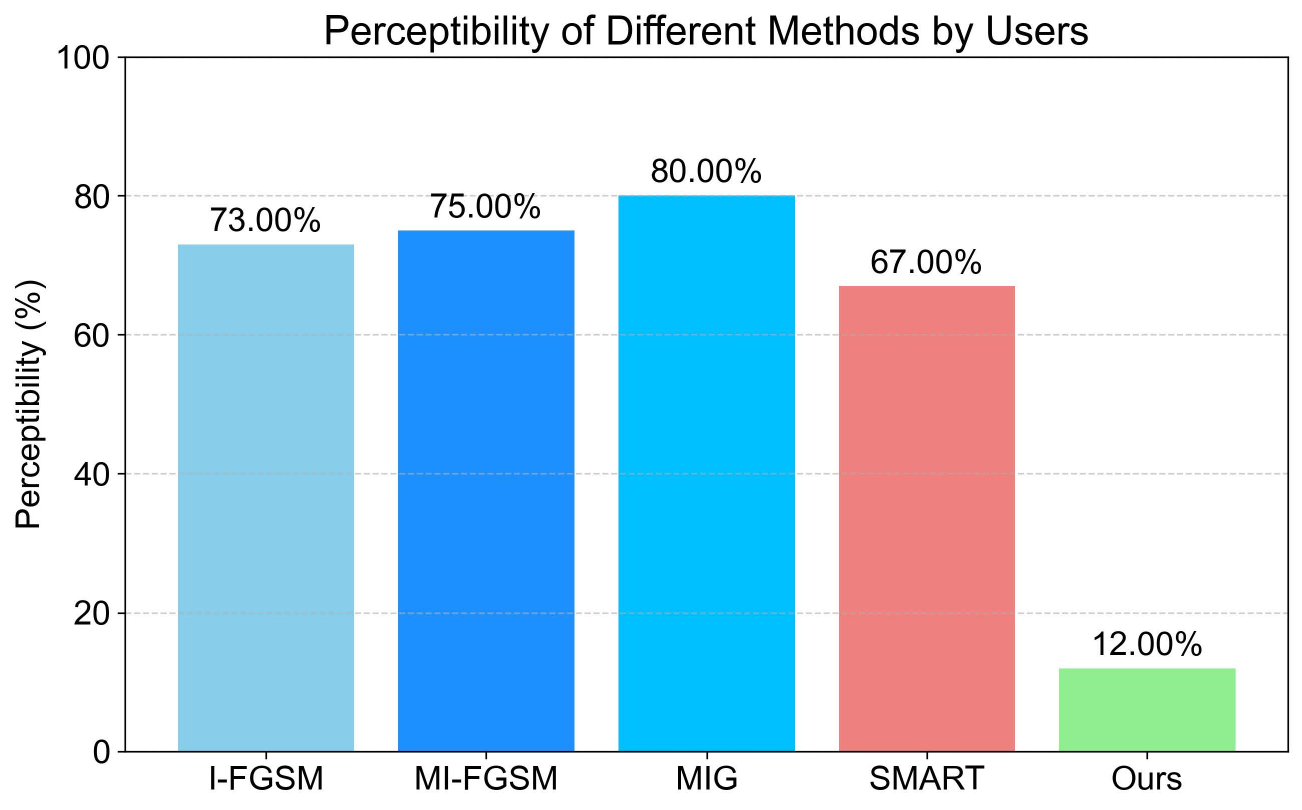}
\caption{Perceptibility comparison across different methods.}
\label{fig:user_study}
\end{figure}

Additionally, we recruited volunteers without visual impairments from diverse backgrounds and a balanced gender representation. They participate in questionnaires to evaluate the imperceptibility of our adversarial samples.
We provided participants with batches of motions from the same label, which pre-attack motions and post-attack motions generated by our method are mixed together.
Participants are asked to select the motions they consider most likely to have potentially been attacked without any time constraints.
As illustrated in Fig.~\ref{fig:user_study}, our method produces adversarial motions that are the least perceivable by humans.

\subsection{Black-Box Settings}

{\color{black}Our method does not require gradient access to the victim classifier during optimization. To further evaluate its effectiveness under restricted-access conditions, we additionally report black-box experiments. For the gradient-based baselines, we adopt a transfer-based black-box protocol: for each victim model, the other three models are used as substitute models in turn, and we report the averaged transfer performance. The same maximum attack budget as in the main paper is used, namely 2000 iterations. By contrast, our method does not rely on surrogate models or victim gradients during optimization. In the current black-box evaluation, the victim model is only checked after the full attack procedure to determine whether the generated adversarial motion succeeds. Randomized smoothing is applied consistently across all reported results. Note that, unlike the transfer-based baselines, our method does not consult the victim model during optimization and therefore cannot exploit intermediate victim-side stopping decisions.} Table \ref{tab:comparison_black_100style} shows the results on 100STYLE dataset in the black-box setting. Our method shows superiority on both deceitfulness and naturalness. Notably, our presented attack method does not involve explicit constraints such as the bone variation while SMART relies on a series of heuristic constraints for the optimization. The experiment shows that our method is versatile to be applied for both white-box and black-box settings without any modification.

\begin{table*}[]
    \caption{Generated Black-box Adversarial Motion Quality Comparison on 100STYLE dataset.}
    \centering
    \begin{tabular}{ccccccc}
    \toprule
    Victim & Method & Success Rate $\uparrow$ & FID $\downarrow$ & MMD $\downarrow$ & Physiological Naturalness $\downarrow$ & Bone Variation $\downarrow$ \\ 
    \toprule
    \multirow{5}{*}{Style \cite{zhong2025smoodi}}
                              & I-FGSM  & 49.46\% & 281.00 & 0.097 & 79.29 & 11.43 \\
                              & MI-FGSM & 51.61\% & 283.70 & 0.098 & 87.51 & 12.61 \\
                              & MIG     & 67.38\% & 307.14 & 0.105 & 414.82 & 66.07 \\
                              & SMART   & 49.46\% & 240.27 & 0.071 & 57.87 & \textbf{4.83} \\
                              & TASAR & 51.97\% & 182.23 & 0.060 & 46.87 & 4.96 \\
                              & Ours    & \textbf{96.42\%} & \textbf{57.50} & \textbf{0.020} & \textbf{21.74} & 5.90 \\ \midrule
    \multirow{5}{*}{STTFormer \cite{qiu2022spatio}}
                              & I-FGSM  & 36.20\% & 194.05 & 0.046 & 62.67 & 11.21 \\
                              & MI-FGSM & 39.07\% & 200.65 & 0.048 & 69.48 & 12.47 \\
                              & MIG     & 65.95\% & 297.48 & 0.104 & 293.88 & 52.50 \\
                              & SMART   & 35.13\% & 193.07 & 0.049 & 57.76 & \textbf{4.80} \\
                              & TASAR & 39.78\% & 162.31 & 0.039 & 43.21 & 4.90 \\
                              & Ours  & \textbf{93.90\%}   & \textbf{62.01} & \textbf{0.023} & \textbf{25.42} &               6.10 \\ \midrule
    \multirow{5}{*}{Skateformer \cite{do2024skateformer}}
                              & I-FGSM  & 41.22\% & 134.05 & 0.036 & 18.77 & 3.70 \\
                              & MI-FGSM & 42.65\% & 138.07 & 0.037 & 20.23 & 4.02 \\
                              & MIG     & 55.20\% & 240.37 & 0.066 & 97.03 & 18.34 \\
                              & SMART   & 39.07\% & 111.17 & 0.028 & 17.74 & \textbf{1.91} \\
                              & TASAR   & 42.30\% & 99.25 & 0.023 & 15.21 & 2.01 \\
                              & Ours    & \textbf{99.64\%}   & \textbf{19.51} & \textbf{0.012} & \textbf{10.97} & 2.86 \\ \midrule
    \multirow{5}{*}{FR-Head \cite{zhou2023learning}}
                              & I-FGSM  & 39.78\% & 275.96 & 0.098 & 111.98 & 16.22 \\
                              & MI-FGSM & 45.52\% & 284.80 & 0.103 & 124.31 & 18.07 \\
                              & MIG     & 70.97\% & 336.28 & 0.156 & 552.29 & 77.84 \\
                              & SMART   & 43.37\% & 266.16 & 0.099 & 123.33 & 8.56 \\
                              & TASAR   & 48.75\% & 243.41 & 0.079 & 104.72 & 7.42 \\
                              & Ours    & \textbf{100\%} & \textbf{20.02} & \textbf{0.012} & \textbf{11.21} & \textbf{3.06}\\ \midrule
    \multirow{5}{*}{\textit{Average}}  
                              & I-FGSM  & 41.67\% & 221.27 & 0.069 & 68.18 & 10.64 \\
                              & MI-FGSM & 44.71\% & 226.81 & 0.072 & 70.38 & 11.79 \\
                              & MIG     & 64.88\% & 295.32 & 0.108 & 339.51 & 53.69 \\
                              & SMART   & 41.76\% & 202.67 & 0.062 & 64.18 & 5.03 \\
                              & TASAR   & 45.70\% & 171.80 & 0.050 & 52.50 & 4.82 \\
                              & Ours    & \textbf{97.49\%} & \textbf{39.76} & \textbf{0.017} & \textbf{17.34} & \textbf{4.48} \\
                              \bottomrule
    \end{tabular}
    \label{tab:comparison_black_100style}
\end{table*}

\begin{table*}[]
\caption{The quality of adversarial motions generated via different variants and configurations.}
\centering
\begin{tabular}{cccccccc}
\toprule
Variants & Configuration & Success Rate $\uparrow$ & FID $\downarrow$ & MMD $\downarrow$ & Physiological Naturalness $\downarrow$ & Foot Skating $\downarrow$ & Bone Variation $\downarrow$ \\ 
\toprule
\multirow{2}{*}{Timestep}     & [1, 20]    & 100\% & 14.01 & 0.007 & 13.28 & 0.090 & 3.22 \\
                              & [980, 1000] & 100\% & 26.14 & 0.017 & 30.37 & 0.050 & 1.62 \\
\midrule
\multirow{2}{*}{Latent}       & $\bm\kappa_t$ & 100\% & 18.91 & 0.011 & 16.05 & 0.075 & 2.94 \\
                              & $\hat{\mathbf{x}}_0$ & 100\% & 19.26 & 0.011 & 56.70 & 0.083 & 3.31 \\ 
\midrule
\multirow{2}{*}{Architecture} & Diffusion \cite{tevet2023human} & 100\% & 18.91 & 0.011 & 16.05 & 0.075 & 2.94 \\
                              & VAE \cite{petrovich2021action} & 96.06\% & 198.00 & 0.056 & 1058.08 & 0.083 & 214.75 \\
\bottomrule
\end{tabular}
\label{tab:ablation}
\end{table*}

\subsection{Ablation Study}

Given that our method is based on the latent space of a diffusion model, we first perform an ablation study to investigate the influence of the chosen timesteps in constructing the stochastic latent space.
Secondly, we examine the impact of different latents used for adversarial attacks to determine which latent is the most suitable for the quality-preserving imperceptible attack task.
Finally, we validate the choice of diffusion models by exploring alternative generative models for quality-preserving adversarial attack.

\subsubsection{Timestep of Stochastic Latent Space}

We investigate the impact of different timestep ranges used to map motions from data space to diffusion latent space for adversarial attacks.
It has been shown that the semantics encapsulated by the latents have smooth transitions with respect to the diffusion timesteps \cite{sclocchi2024phase,huang2023dreamtime}, we conduct experiments using two representative timestep ranges: the earliest timestep range ($t \sim [1, 20]$) and the latest timestep range ($t \sim [980, 1000]$) for comparison.

A trade-off exists between motion naturalness and plausibility, as shown in Table \ref{tab:ablation}.
This trade-off effect arises from the different emphasis on semantics encoded in the latents.
Generally, latents derived from earlier timesteps primarily capture low-level details, while those from later timesteps focus on high-level structural patterns \cite{kwon2023diffusion}.
Utilizing earlier timesteps to construct the stochastic distributional latents is biased toward detailed movements, with a reduced understanding of global coherency and consistency.
This bias enhances naturalness by conveying motion dynamics.
However, it also exacerbates foot skating and bone length variation due to the lack of global rationality in the latents.
Conversely, constructing the distributional latents using later timesteps undermines naturalness, as the generated adversarial motions deviate from the natural distribution, leading to higher FID and MMD scores.

\subsubsection{Alternative Latents for Attack}

We examine the influence of different latents used for our quality-preserving adversarial attacks.
Specifically, we compare the constructed latent with the predicted pre-attack data as the latent for adversarial attack.
The constructed latent $\bm\kappa$ evaluates the required perturbations based on a single timestep of denoising with regularization, and allows for step-by-step motion modifications.
In contrast, the predicted pre-attack data $\hat{\mathbf{x}}_0$ represents modifications along the entire chain of previous timesteps with approximation.
We conduct experiments to determine whether the constructed latent is more effective to preserve post-attack motion quality than leveraging the predicted pre-attack data.

As shown in Table \ref{tab:ablation}, using the predicted pre-attack data as the latent leads to a consistent decline in all performances of adversarial attack.
Although both latent choices achieve the same success rate, the quality of adversarial motions deteriorates significantly when switching from the constructed latent to the predicted data.
This deterioration stems from the approximation error when calculating the required latents.
The predicted $\hat{\mathbf{x}}_0$ is derived from the $\mathbf{x}_{t+1}$ by approximating all previous timesteps $\{i\}_{i=1}^t$ collectively, thereby ignoring information from intermediate timesteps.
Conversely, the constructed latent $\kappa_t$, also derived from the $\mathbf{x}_{t+1}$, considers only the desired changes within a single timestep.
This finer-grained modification enhances the quality preservation during adversarial attacks by integrating fewer approximation errors.

\subsubsection{Alternative Distribution Modeller}

Given that our attack method is based on generative models, we explore the use of alternative generative architectures. Similarly, we utilize their latent spaces to modify motions adversarially.
Specifically, we conduct experiments using either diffusion models or variational autoencoders (VAEs). We train the VAE with label conditions following the architecture and training configurations outlined in \cite{petrovich2021action}. To maintain consistency with our use of the latent distribution, we employ the latent distribution in the VAE. All other configurations remain consistent with those used in our diffusion-based method.

As shown in Table \ref{tab:ablation}, the performance of adversarial attacks using a VAE model significantly declines compared to our diffusion-based method. It indicates that effective motion modification requires a comprehensive representation of underlying patterns. In contrast to diffusion models, VAEs utilize a single latent feature rather than a hierarchy of features. Consequently, using a single latent feature in VAEs results in ambiguity regarding motion dynamics and leads to under-expressed movement coherency and consistency.

\begin{table*}[htbp]
\caption{{\color{black}Generated Adversarial Motion Quality Comparison on 100STYLE dataset against Style \cite{zhong2025smoodi}. Expectation over Transformation (EoT)~\cite{athalye2018synthesizing} has been integrated on I-FGSM, MI-FGSM, MIG, and SMART.}}
\centering
\color{black}
\begin{tabular}{ccccccc}
\toprule
Method & Success Rate $\uparrow$ & FID $\downarrow$ & MMD $\downarrow$ & Physiological Naturalness $\downarrow$ & Foot Skating $\downarrow$ & Bone Variation $\downarrow$ \\
\midrule
I-FGSM & 81.72\% & 185.09 & 0.051 & 117.48 & 0.129 & 9.91 \\
MI-FGSM & 81.72\% & 185.58 & 0.052 & 116.92 & 0.128 & 9.86 \\
MIG & 70.97\% & 241.43 & 0.066 & 281.46 & 0.222 & 24.96 \\
SMART & 81.00\% & 247.50 & 0.079 & 341.36 & 0.306 & 32.63 \\
Ours   & \textbf{100\%} & \textbf{18.91} & \textbf{0.011} & \textbf{16.05}  & \textbf{0.075} & \textbf{2.94}  \\
\bottomrule
\end{tabular}
\label{tab:adaptive_attack}
\end{table*}

{\color{black}\subsubsection{Adaptive Strategy}
Table~\ref{tab:adaptive_attack} reports the results after equipping the gradient-based baselines with EoT under the randomized smoothing setting. Compared with the original comparison, adaptive gradient estimation improves the effectiveness of some baselines, most notably SMART, whose success rate increases substantially to 81.00\%, indicating that part of its previous performance degradation indeed came from the stochastic defense rather than the attack formulation itself. However, even after this stronger adaptive evaluation, our method still consistently outperforms all baselines by a large margin in both attack success and post-attack motion quality. In particular, our method achieves a 100\% success rate while maintaining dramatically lower FID, MMD, physiological naturalness, foot skating, and bone variation, showing that the advantage of our approach does not rely on a non-adaptive evaluation protocol. These results suggest that although EoT can partially recover the effectiveness of gradient-based attacks against randomized models, such methods still tend to introduce noticeable quality degradation, whereas our distribution-based attack remains substantially more effective at preserving the naturalness and plausibility of adversarial motions. A possible explanation is that EoT makes the estimated gradient more consistent with the randomized victim, which improves attack success, but it does not alter the fact that these baselines still optimize adversarial perturbations in the original sample space. As a result, the recovered adversarial strength comes at the expense of larger motion distortion, whereas our method constrains the attack through the learned motion distribution and therefore preserves motion quality much more effectively.
}

{\color{black}\subsubsection{Time Cost Analysis}

\begin{table}[htbp]
\caption{{\color{black}Runtime comparison between our method and I-FGSM. We report the average runtime per attack and the average runtime per iteration.}}
\centering
\color{black}
\begin{tabular}{lcc}
\toprule
Method & Per attack (s) & Per iteration (s) \\
\midrule
I-FGSM & 18.05 & 0.06 \\
Ours   & 1.72  & 0.07 \\
\bottomrule
\end{tabular}
\label{tab:time_cost}
\end{table}

We further evaluate the computational efficiency of different attack methods by reporting two runtime statistics. The first is the average runtime per attack, computed as the total time of all attack attempts divided by the number of attacks, which reflects the effective time cost required to obtain one adversarial motion. The second is the average runtime per iteration, which measures the computational overhead of a single optimization step. As shown in Table \ref{tab:time_cost}, our method requires 1.72 seconds per attack on average, compared with 18.05 seconds for I-FGSM, while the per-iteration runtime is 0.07 seconds for our method and 0.06 seconds for I-FGSM. These results show that the additional denoising module introduces only a marginal per-iteration overhead, whereas the overall runtime is substantially reduced in terms of attack generation. The iteration speed is comparable because our method does not rely on iterative gradient back-propagation through the victim classifier and only performs a single-step denoising computation over two adjacent diffusion timesteps rather than traversing the full diffusion chain.
}

\section{Conclusion}\label{sec:conclusion_discussion}

We propose a novel attack application that imperceptible adversarial motions are achieved without compromising post-attack motion quality.
Additionally, we introduce a distribution-based adversarial attack method targeting skeleton-based human action recognition (S-HAR) systems by minimizing the optimization gap inherent in previous approaches.
Our method integrates a generative diffusion model, wherein the posterior mean of single timestep denoising with regularization is constructed as the proxy to fulfill our attack strategy.
To faithfully assess the naturalness of adversarial motions, we develop a new metric aligned with human perception of natural, real-world human movements.
We evaluate the quality of adversarial motions in terms of threatfulness, motion quality, and imperceptibility, demonstrating that our adversarial motions achieve superior performance across these metrics.
The success of our proposed quality-preserving attack application and distribution-based attack method raises significant concerns regarding the robustness of action recognizers, highlighting the necessity for further enhancements in this area.

While achieving natural adversarial motions, our approach presents opportunities for future research. As shown in our experiment, there exists a trade-off in motion quality with respect to the timesteps of the diffusion model. Future work can potentially investigate the influence of trading-off post-attack motion quality on different S-HAR systems. {\color{black} It is also valuable to investigate noise-aware or disentangled diffusion priors that explicitly separate motion dynamics from acquisition noise, especially for lower-fidelity datasets where the learned data distribution may itself contain substantial noise.} Our proposed physiological naturalness focuses on the physiological perspective of motion naturalness. Future work can potentially integrate the research field of action quality assessment \cite{zhou2024magr,zhou2023video,zhou25phi} and character animation \cite{wang2019spatio} for more comprehensive motion quality measurements as well as constraints.

\begin{IEEEbiography}
[{\includegraphics[width=1in,height=1.1in,clip,keepaspectratio]{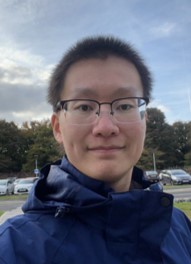}}]{Ziyi Chang}
is a PhD student in the Department of Computer Science at Durham University. His research focuses on diffusion models with human motions. Specifically, his research involves diffusion models for styled skeleton-based human motion synthesis, skeleton-based human motion analysis, skeleton-based human interaction modelling. He also has interest in 3D surface reconstruction and domain adaptation. He received MSc degree from the University of Edinburgh in 2020 and BSc degree from Renmin University of China in 2019.
\end{IEEEbiography}

\vspace{-20pt}

\begin{IEEEbiography}[{\includegraphics[width=1in,height=1.1in,clip,keepaspectratio]{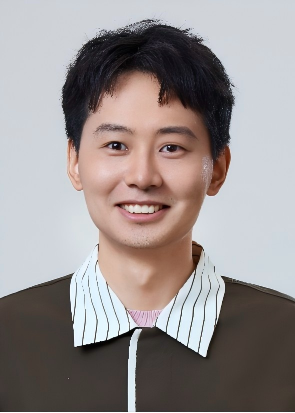}}]{Kanglei Zhou} 
is currently a Postdoctoral Researcher with the Department of Psychology and Cognitive Science, Tsinghua University. He received his Ph.D. degree in Computer Science and Engineering from Beihang University in 2025 and his B.E. degree from Henan Normal University in 2020. In 2024, he was a Visiting Student at Durham University. His research interests include human motion analysis and continual learning.
\end{IEEEbiography}

\vspace{-20pt}

\begin{IEEEbiography}[{\includegraphics[width=1in,height=1.1in,clip,keepaspectratio]{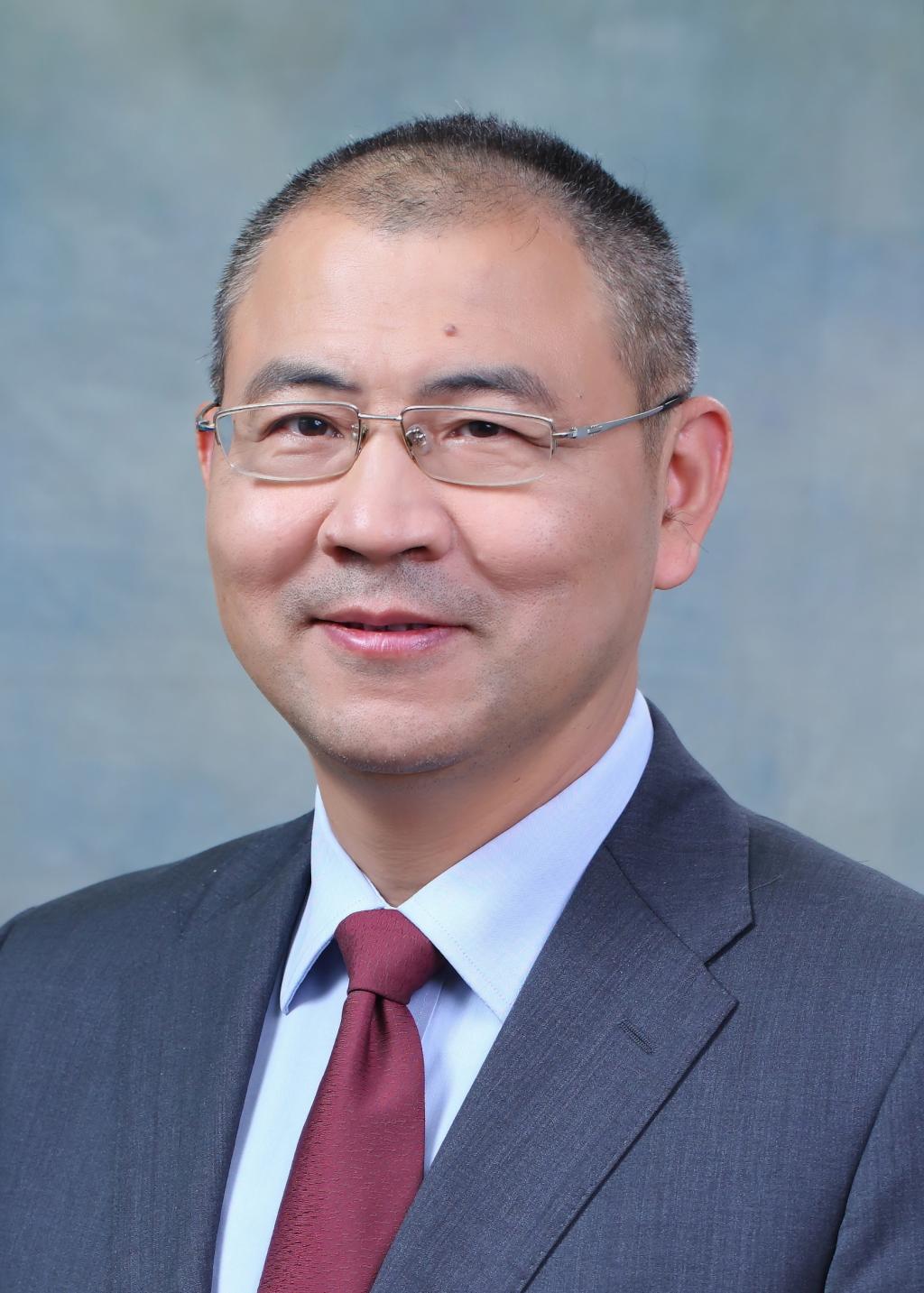}}]{Xiaohui Liang} received his Ph.D. degree in computer science and engineering from Beihang University, China. He is currently a Professor, working in the School of Computer Science and Engineering at Beihang University. His main research interests
include computer graphics and animation, visualization, and virtual reality.
\end{IEEEbiography}

\vspace{-20pt}

\begin{IEEEbiography}[{\includegraphics[width=1in,height=1.1in,clip,keepaspectratio]{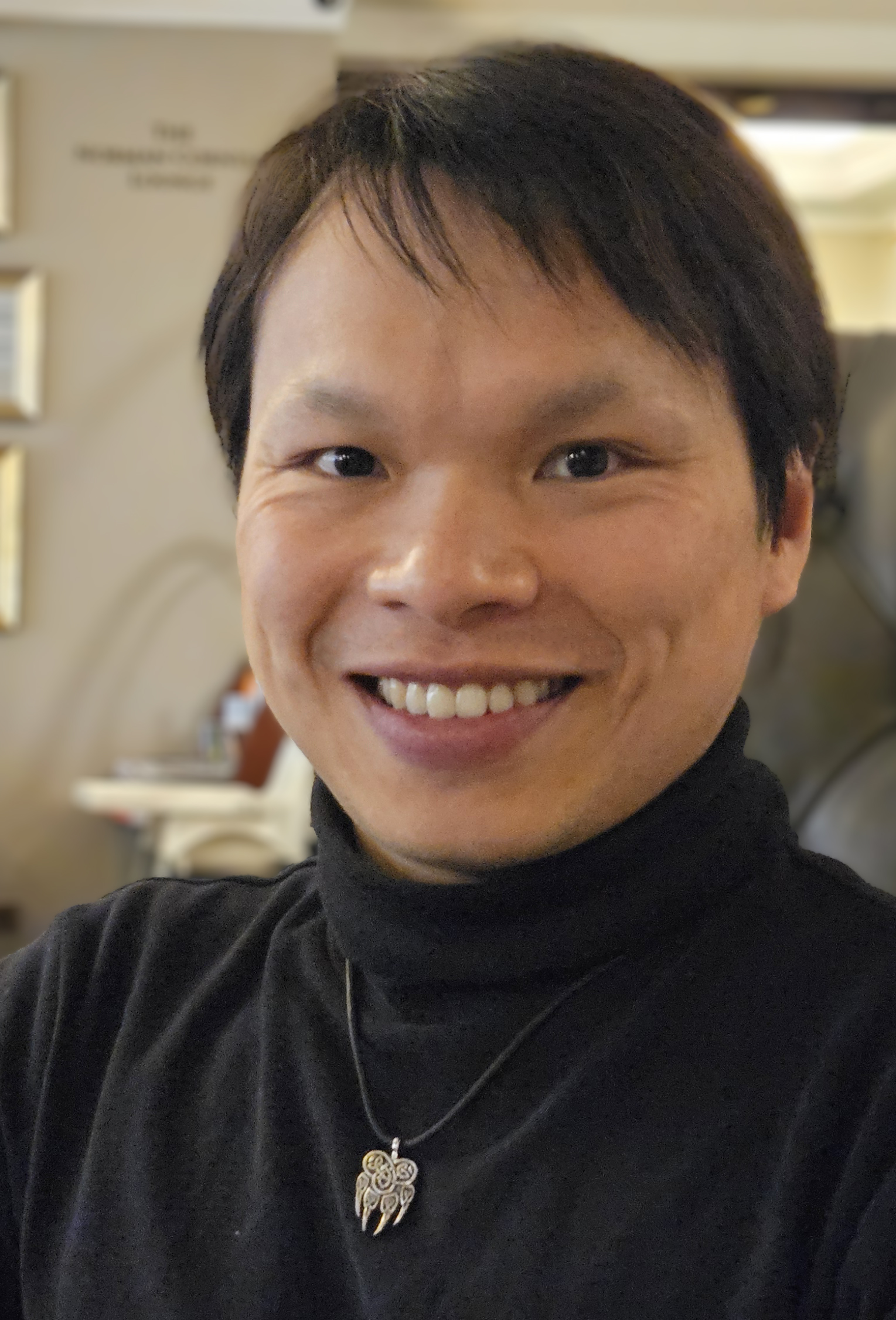}}]{Hubert P. H. Shum}
(Senior Member, IEEE) is a Professor of Visual Computing and the Director of Research in the Department of Computer Science at Durham University, specialising in modelling spatio-temporal information with responsible AI. He is also a Co-Founder and the Co-Director of Durham University Space Research Centre. Before this, he was an Associate Professor at Northumbria University and a Postdoctoral Researcher at RIKEN Japan. He received his PhD degree from the University of Edinburgh. He chaired conferences such as Pacific Graphics, BMVC and SCA. He has authored over 200 research publications in the fields of Computer Vision, Computer Graphics and AI in Healthcare, underpinned by Responsible AI designs and algorithms.
\end{IEEEbiography}

\vfill

\end{document}